\documentclass[10pt,journal,compsoc]{IEEEtran}

\ifCLASSOPTIONcompsoc
  \usepackage[nocompress]{cite}
\else
  \usepackage{cite}
\fi
\ifCLASSINFOpdf
\else
\fi
\pdfoutput=1

\usepackage{url}
\usepackage{times}
\usepackage{epsfig}
\usepackage{graphicx}
\usepackage{amsmath}
\usepackage{amssymb}
\usepackage{multirow}
\usepackage{colortbl}
\usepackage{color}
\usepackage{threeparttable}
\usepackage{booktabs}
\usepackage{adjustbox}
\usepackage{caption}
\usepackage{subcaption}
\usepackage[misc]{ifsym}
\usepackage{xcolor,colortbl}
\usepackage{makecell}
\newcommand*{\belowrulesepcolor}[1]{%
  \noalign{%
    \kern-\belowrulesep 
    \begingroup 
      \color{#1}%
      \hrule height\belowrulesep 
    \endgroup 
    \vspace{-0.03mm}
  }%
} 
\newcommand*{\aboverulesepcolor}[1]{%
  \noalign{%
  \vspace{-0.03mm}
    \begingroup 
      \color{#1}%
      \hrule height\aboverulesep 
    \endgroup 
    \kern-\aboverulesep 
  }%
}
\newcommand{\vspacefigtext}{\vspace{-3mm}}

\usepackage{pifont}
\newcommand{\cmark}{\ding{51}}%
\newcommand{\xmark}{\ding{55}}%

\usepackage{xspace}
\makeatletter
\DeclareRobustCommand\onedot{\futurelet\@let@token\@onedot}
\def\@onedot{\ifx\@let@token.\else.\null\fi\xspace}

\usepackage[linesnumbered,lined,boxed,commentsnumbered,ruled]{algorithm2e}
\usepackage[pagebackref,breaklinks=true,colorlinks,citecolor=blue,urlcolor=blue,linkcolor=blue,bookmarks=false]{hyperref}
\begin{document}
%

\title{Towards Open Vocabulary Learning: \\ A Survey}

%
%
%
%

\author{ 
        Jianzong Wu*,
        Xiangtai Li* \textsuperscript{$\dagger$},
        Shilin Xu*,
        Haobo Yuan*,
        Henghui Ding,
        Yibo Yang, 
        Xia Li, \\
        Jiangning Zhang,
        Yunhai Tong, 
        Xudong Jiang,
        Bernard Ghanem,
        Dacheng Tao
\IEEEcompsocitemizethanks{
\IEEEcompsocthanksitem J.~Wu, S.~Xu, X.~Li, and Y.~Tong are with the National Key Laboratory of General Artificial Intelligence, School of Intelligence Science and Technology, Peking University, Beijing, China. Corresponding Author \textsuperscript{$\dagger$}: Xiangtai Li. lxtpku@pku.edu.cn
\IEEEcompsocthanksitem H.~Ding, X.~Jiang are with Nanyang Technological University, Singapore. 
\IEEEcompsocthanksitem H.~Yuan is with Wuhan University, Wuhan, China.
\IEEEcompsocthanksitem Y. Yang and B. Ghanem are with King Abdullah University of Science and Technology, Saudi Arabia.
\IEEEcompsocthanksitem D.~Tao is with the University of Sydney, Sydney, Australia.
\IEEEcompsocthanksitem X.~Li is with ETH Zurich, Switzerland.
\IEEEcompsocthanksitem J.~Zhang is with Zhejiang University, Hangzhou, China.
\IEEEcompsocthanksitem The \textbf{first four authors} share the same technical contribution to this work. 
}
}

%
%

\markboth{IEEE TRANSACTIONS ON PATTERN ANALYSIS AND MACHINE INTELLIGENCE}
{Shell \MakeLowercase{\textit{et al.}}: Bare Advanced Demo of IEEEtran.cls for IEEE Computer Society Journals}
%



\IEEEtitleabstractindextext{
\begin{abstract}
In the field of visual scene understanding, deep neural networks have made impressive advancements in various core tasks like segmentation, tracking, and detection. However, most approaches operate on the close-set assumption, meaning that the model can only identify pre-defined categories that are present in the training set. Recently, open vocabulary settings were proposed due to the rapid progress of vision language pre-training. These new approaches seek to locate and recognize categories beyond the annotated label space. The open vocabulary approach is more general, practical, and effective than weakly supervised and zero-shot settings.
This paper thoroughly reviews open vocabulary learning, summarizing and analyzing recent developments in the field. In particular, we begin by juxtaposing open vocabulary learning with analogous concepts such as zero-shot learning, open-set recognition, and out-of-distribution detection. Subsequently, we examine several pertinent tasks within the realms of segmentation and detection, encompassing long-tail problems, few-shot, and zero-shot settings.
As a foundation for our method survey, we first elucidate the fundamental principles of detection and segmentation in close-set scenarios.
Next, we examine various contexts where open vocabulary learning is employed, pinpointing recurring design elements and central themes. 
This is followed by a comparative analysis of recent detection and segmentation methodologies in commonly used datasets and benchmarks. 
Our review culminates with a synthesis of insights, challenges, and discourse on prospective research trajectories. To our knowledge, this constitutes the inaugural exhaustive literature review on open vocabulary learning. We keep tracing related works at \url{https://github.com/jianzongwu/Awesome-Open-Vocabulary}.
\end{abstract}

}

\maketitle

\IEEEdisplaynontitleabstractindextext

%
\IEEEpeerreviewmaketitle

\section{Introduction}


\IEEEPARstart{D}{eep} neural networks have revolutionized scene understanding tasks, including object detection, segmentation, and tracking~\cite{resnet,maskrcnn,VIT,detr}. 
Nonetheless, using conventional approaches for real-world applications can be challenging due to limitations such as inadequate class annotations, close-set class definitions, and costly labeling expenses. 
These limitations can increase the difficulty and cost of implementing a deep model on new scenes, particularly when the number of categories or concepts in the scene is much larger than what is included in the training dataset.

For example, object detection is a core computer vision task involving scene understanding. It requires human annotations for each category and each object location, which can be costly and time-consuming. 
For instance, the COCO dataset~\cite{COCO_dataset}, widely used for benchmarking object detection algorithms, only includes 80 categories. 
But in reality, natural scene images often have more than 80 different types of objects. 
We would need to incur significant annotation costs to extend object detectors to cover all these categories. 
Current research focuses on developing methods to train more flexible object detectors on the subset of COCO with base classes and let them identify new or unfamiliar objects without requiring additional annotations.

Several previous solutions adopt zero-shot learning (ZSL)~\cite{romera2015embarrassingly,bansal2018zero,xian2016latent}. 
These approaches extend a detector to generalize from annotated (seen) object classes to other (unseen) categories. 
The annotations of seen object classes are used during the training, while the annotations of unseen classes are \textit{strictly} unavailable during training. 
Most approaches adopt word embedding projection to constitute the classifier for unseen class classification. 

However, these approaches come with several limitations. ZSL, in particular, is highly restrictive. 
Typically, these methods lack examples of unseen objects and treat these objects as background objects during training. 
As a result, during inference, the model identifies novel classes solely based on their pre-defined word embeddings~\cite{devlin2018bert,mikolov2013distributed_word2vec}, thereby limiting exploration of the visual information and relationships of those unseen classes. This is why ZSL approaches have been shown to yield unsatisfying results in novel classes.

Open vocabulary learning is proposed to handle the above issue and make the detector extendable. It has been successfully applied to multiple tasks, e.g., segmentation, detection, video understanding, scene understanding, etc. In particular, open vocabulary object detection~\cite{zareian2021opendet_ovrcnn,gu2021open_vild} is first proposed. Also, open vocabulary segmentation is proposed~\cite{ghiasi2021openvoc_seg, Language-driven-semantic-segmentation}. Similarly, the model trains on base classes and inferences on both base and novel classes. The critical difference between zero-shot learning and open vocabulary is that one can use \textit{visual-related language vocabulary data} like image captions as auxiliary supervision in open vocabulary settings. The motivations to use language data as auxiliary weak supervision are: 
1) Language data requires less labeling effort and thus is more cost-friendly. 
The visual-related language data, like image captions, is widely available and more affordable than box or mask annotations.
2) Language data provides a more extensive vocabulary size and thus is more extendable and general. For example, words in captions are not limited to the pre-defined base categories. It may contain novel class names, attributes, and motions of objects. Incorporating captions in training has been proven to be extremely useful in helping improve the models' scalability.

Moreover, recently, visual language models (VLMs)~\cite{CLIP,jia2021scaling_align}, which pre-train themselves on large-scale image-text pairs, show remarkable zero-shot performance on various vision tasks. 
The VLMs align images and language vocabularies into the same feature space, fulfilling the visual and language data gap. 
Many open vocabulary methods effectively eliminate the distinction between close-set and open-set scenarios by utilizing the alignment learned in VLMs, making them highly suitable for practical applications. 
For example, An open vocabulary object detector can be easily extended to other domains according to the demand without the need to gather relevant data or incur additional labeling expenses.


As open vocabulary models continue to advance rapidly and demonstrate impressive results, it is worthwhile to track and compare recent research on open vocabulary learning.
Several surveys work on low-shot learning~\cite{song2022comprehensive,wang2020generalizing,wang2019survey,kohler2021few,hu2023suppressing}, including few-shot learning and zero-shot learning. There are also several surveys on multi-modal learning~\cite{xu2022multimodal,baltruvsaitis2018multimodal}, including using transformer for vision language tasks~\cite{li2023transformer,xu2022multimodal} and vision language pre-training~\cite{ruan2022survey}.
However, these works focus on learning with few examples, multi-modal fusion, or pre-training for better feature representation.
As far as we know, there haven't been any surveys that thoroughly summarize the latest developments in open vocabulary learning, including methods, settings, benchmarks, and the use of vision foundation models. We aim to fulfill the blank with this work.

\noindent
\textbf{Contribution.} In this survey, we systematically track and summarize recent literature on open vocabulary learning, including object detection, segmentation, video understanding, and 3D scene understanding. 
The survey covers the most representative works in each domain by extracting the common technical details. 
It also contains the background of open vocabulary learning and related concepts comparison, including zero-shot learning (ZSL)~\cite{zeroshotobjectdetection1}, open set recognition (OSR)~\cite{openworld,sun2021m2iosr}, and out-of-distribution detection (OOD)~\cite{yang2021oodsurvey}. 
It also includes large-scale visual language models and representative detection and segmentation works, which makes the survey self-contained. 
In addition, we present a comprehensive analysis and comparison of benchmarks and settings for each specific domain. As far as we know, we are the first to concentrate on the specific area.
Finally, since the field of open vocabulary learning is rapidly evolving, we may not be able to keep up with all the latest developments. We welcome researchers to contact us and share their new findings in this area to keep us updated. Those new works will be included and discussed in the revised version. 
%

\noindent
\textbf{Survey pipeline.} In Sec.~\ref{sec:background}, we will cover the background knowledge, including definition, datasets, metrics, and related research domains. 
Then, in Sec.~\ref{sec:method_survey}, we conduct main reviews on various methods according to different tasks. 
In particular, we will first include the preliminary knowledge of close-set detection and segmentation methods. 
Then, we present the method details of each direction, including detection, segmentation, video understanding, and 3D scene understanding. 
Next, we point out challenges and future directions in Sec.~\ref{sec:future_direction} and conclude Sec.~\ref{sec:conclusion}.
Finally, the appendix compares the results for different tasks and benchmarks. 
\begin{figure}[!t]
	\centering
	\includegraphics[width=1\linewidth]{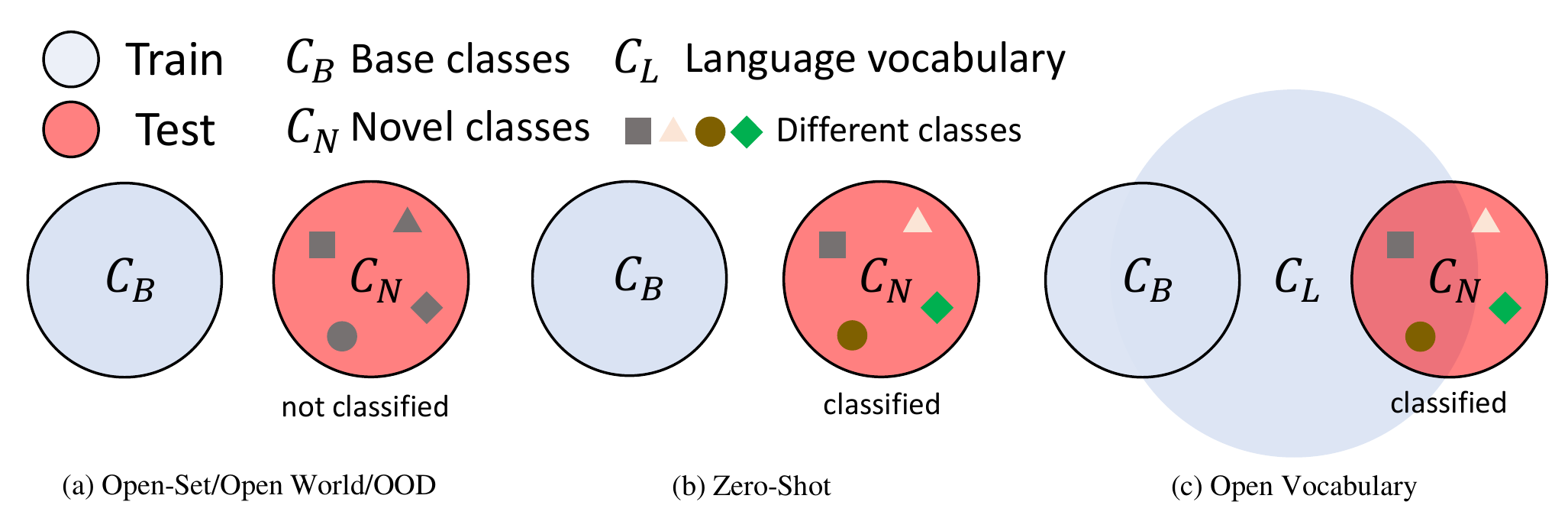}
        \caption{Concepts comparison between open-set/open world/out-of-distribution detection (OOD), zero-shot and open vocabulary. Different shapes represent different novel categories. Colors represent the predictions of the novel objects. (a), in the open-set/open World/OOD settings, the model only needs to identify novel classes and mark them as ``unknown''. (b), in the zero-shot setting, a model must classify unknown classes into specific categories. (c), in the open vocabulary settings, the model can classify novel classes with the help of large language vocabulary knowledge $C_L$.}
	\label{fig:bg-task-diff}
\end{figure}

\section{Background}
\label{sec:background}

\noindent
\textbf{Overview.}
In this section, we first present the concept definition of open vocabulary and related concepts comparison. 
Then, we present a historical review of open vocabulary learning and point out several representatives. 
Next, we present the standard datasets and metrics. We also present the unified notations for open vocabulary object detection and segmentation tasks. 
Finally, we review the related research domains.

\begin{figure}[!t]
	\centering
	\includegraphics[width=1.0\linewidth]{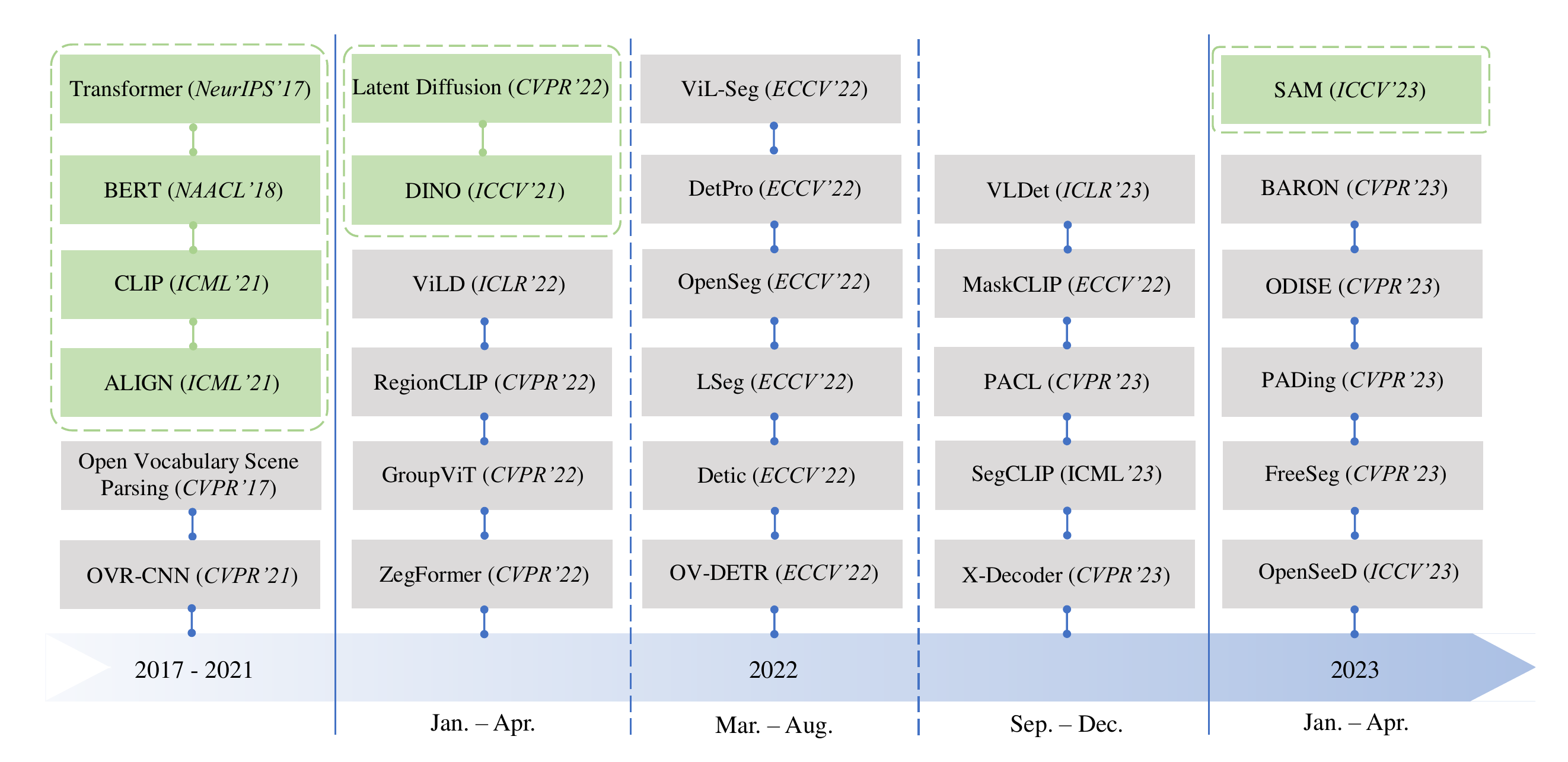}
        \caption{Timeline of open vocabulary learning. The gray boxes indicate representative works. Green boxes indicate the foundation models and VLMs. In open vocabulary learning, many works exploit the knowledge learned by pre-trained vision foundation models like Swin~\cite{liu2021swin} and VLMs like CLIP~\cite{CLIP}. Recently, some works also explore the use of diffusion models in this setting.}
	\label{fig:timeline}
\end{figure}


\begin{figure*}[t!]
    \centering
    \begin{subfigure}{0.3\textwidth}
        \includegraphics[width=\textwidth]{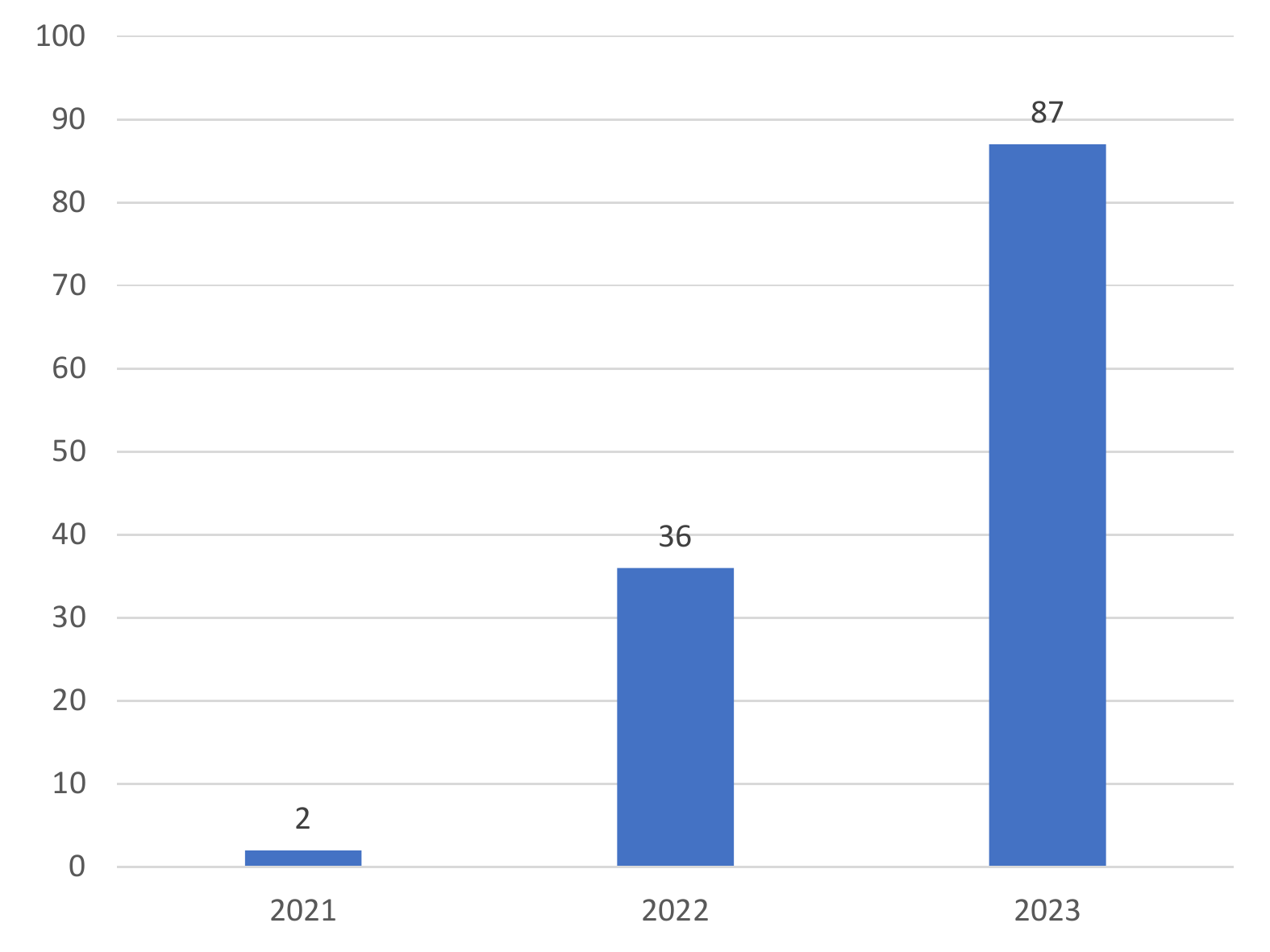}
        \caption{Number of Research Works}
        \label{fig:sub1}
    \end{subfigure}
    \hfill
    \begin{subfigure}{0.26\textwidth}
        \includegraphics[width=\textwidth]{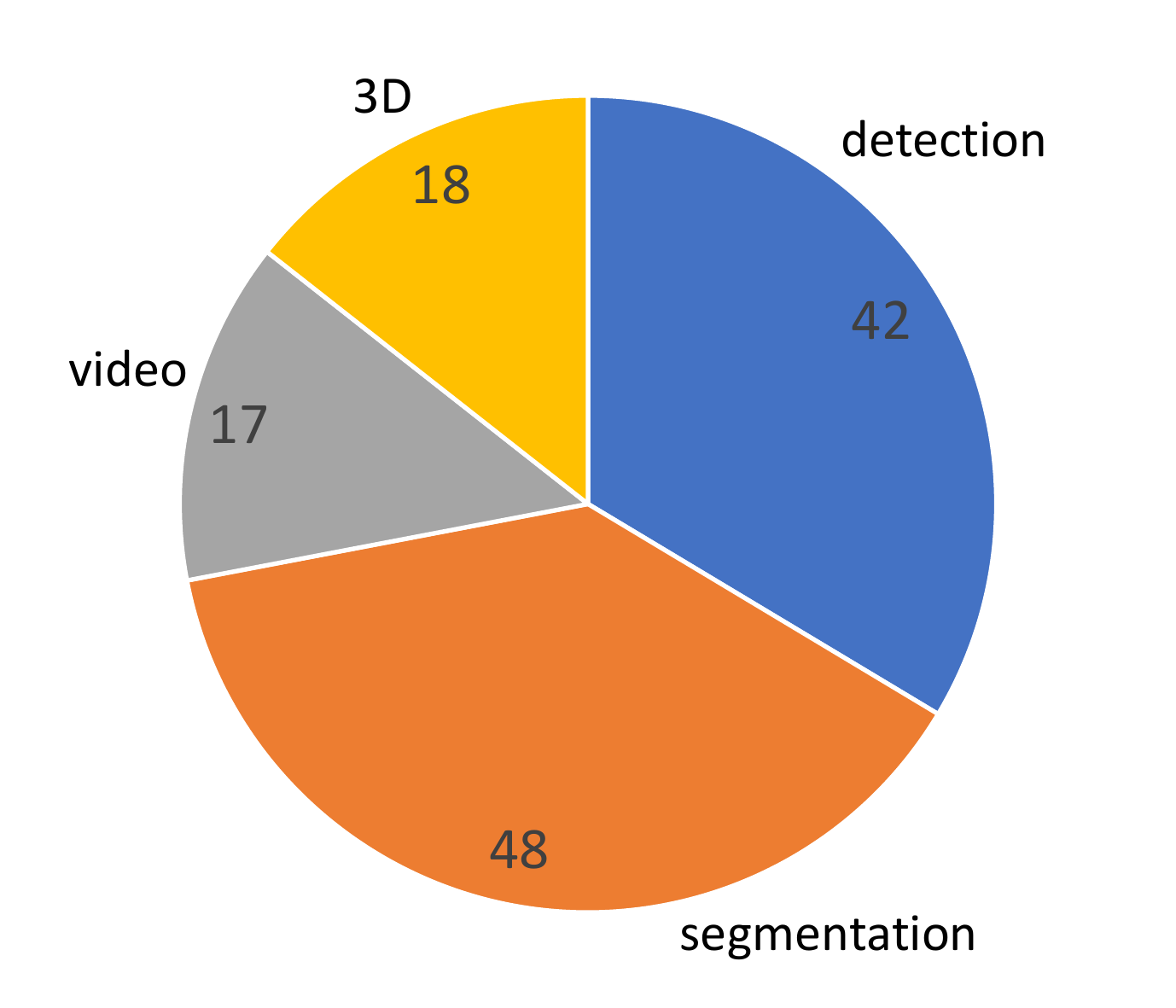}
        \caption{Ratio of Different Directions}
        \label{fig:sub2}
    \end{subfigure}
    \hfill
    \begin{subfigure}{0.3\textwidth}
        \includegraphics[width=\textwidth]{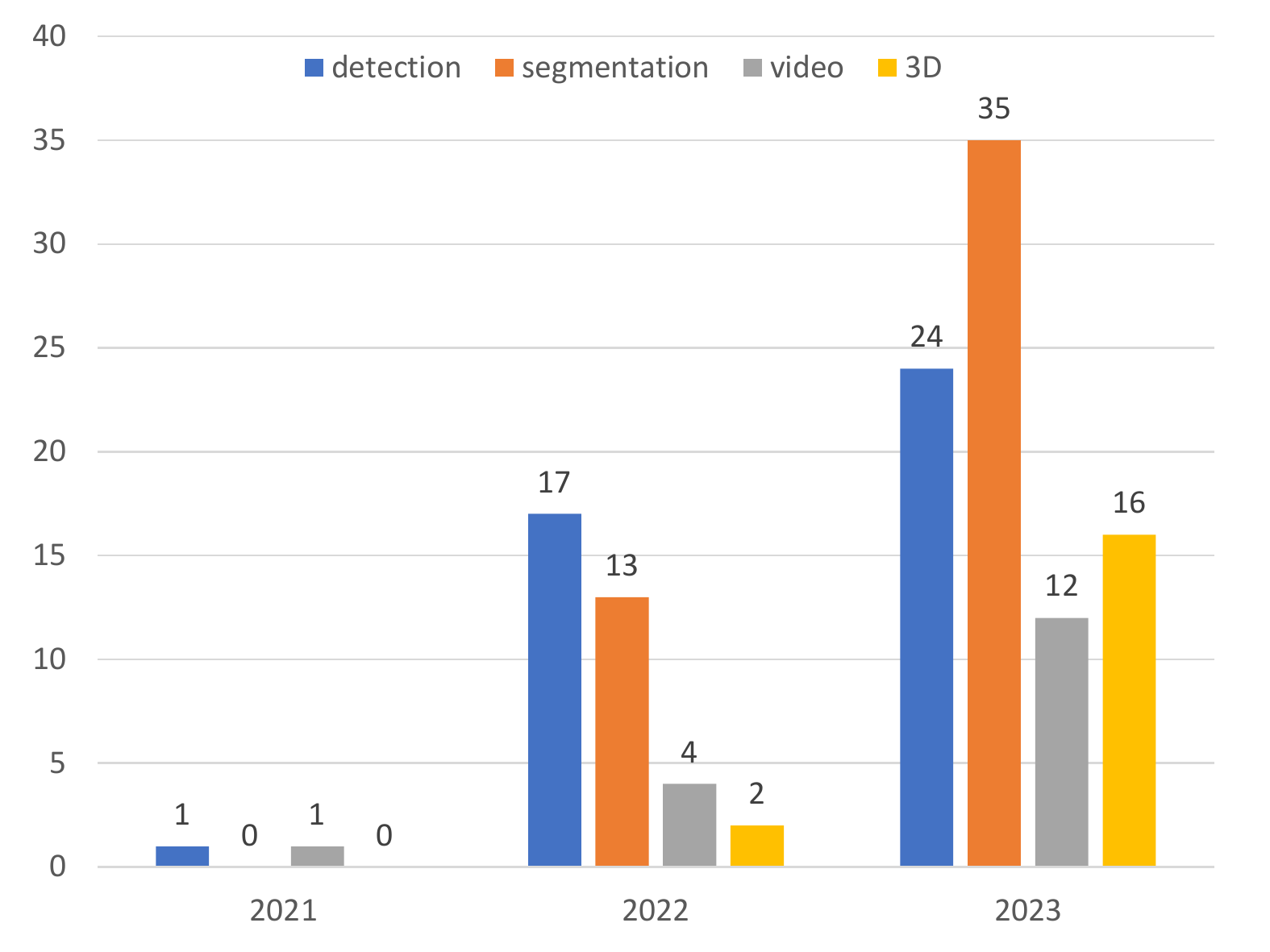}
        \caption{Different Directions per Year}
        \label{fig:sub3}
    \end{subfigure}
    \caption{Summarization on open vocabulary learning works. (a) The number of research works still increases per year. (b) Detection and segmentation have more papers than 3D and video. (c) indicates each direction per year. The results are obtained on 2024/1/15.}
    \label{fig:statistic}
\end{figure*}

\subsection{Concept Definition}

We take the classification task for concept illustration. 
The supervised methods assume that the training data and testing data share the same closed-set label space. However, the model trained under this assumption cannot be extended to new categories. 
To address this issue, researchers have introduced several new concepts like open-set learning and zero-shot learning, ultimately leading to open vocabulary learning.

These concepts have similar settings and notations. In particular, the examples are classified into base and novel classes (or called out-of-distribution examples). 
The base classes can be accessed for training, while the novel classes are not. We denote the base classes in the label set space $C_B$ and the novel classes in the label set space $C_N$.

Due to the low performance on novel classes in earlier works for open-set, open world, and OOD tasks~\cite{open-set-survey-recenet,openworld-detection,yang2021oodsurvey,zareian2021opendet_ovrcnn} and the easier acquisition of image-text pairs, recent works~\cite{zareian2021opendet_ovrcnn,gu2021open_vild} propose the open vocabulary setting. 
It allows using additional low-cost training data~\cite{coco-captions} or pre-trained vision language models like CLIP~\cite{CLIP}, which have much larger language vocabularies $C_L$. 
It may contain the concepts of both $C_B$ and $C_N$, but its main goal is to enable the model to generalize across more classes in the open domain.
In mathematical terms and data view, \textbf{open vocabulary learning} can be defined as follows:

\textbf{Training Data} ($D_{train}$): The training dataset is a collection of data-label pairs, where each pair consists of an input $x_i$ and its associated label $y_i$. The input $x_i$ can be an image or a video according to the task type, and in scene understanding tasks such as detection and segmentation, the label $y_i$ also includes visual labels, such as bounding boxes or masks, in addition to the class labels. Alongside the standard input-label pairs, open vocabulary learning includes vision-aware language vocabulary data, represented as $l_i$. 
The language vocabulary data can be image-caption data or vision-aware class name embeddings in VLMs like CLIP~\cite{CLIP}. The language augmented training set can be denoted as $D_{train} = \{(x_1, y_1, l_1), (x_2, y_2, l_2), ..., (x_n, y_n, l_n)\}$, where $n$ is the training data length, $x_i$ is the visual data, $y_i$ is the label from base classes $C_B$, and $l_i$ is the associated language data from a large vocabulary space $C_L$. 
Note that $C_L$ is not strictly required to contain $C_B$ or $C_N$, as the language vocabulary may not cover all the class names in the vision data. 
On the contrary, $C_L$ may also have words out of the pre-defined novel categories, which can further extend models' generalizability.

\textbf{Evaluation Data} ($D_{eval}$): The evaluation dataset is similarly a collection of data pairs. However, the labels for evaluation data include both base classes and novel classes. This can be represented as $D_{eval} = \{(x'_1, y'_1), (x'_2, y'_2), ..., (x'_m, y'_m)\}$, where $m$ is the evaluation data length, $y'_i$ belongs to either $C_B$ or $C_N$. During the evaluation, open vocabulary methods need to predict $y'_i$ given $x'_i$ in the realm $y'_i \in (C_B \cup C_N)$.

Here, we emphasize the concept differences between open vocabulary learning and analogous concepts as follows:

\noindent$\bullet$
\textbf{Open-Set Learning.}
Open-set learning aims to classify known classes and reject unknown classes during testing~\cite{open-set-survey-recenet,open-set-dis-1,open-set-dis-2}. Concretely, the training data is $D_{train} = \{(x_1, y_1), (x_2, y_2), ..., (x_n, y_n)\}$, where $y_i$ is from base classes $C_B$. 
These are also called known classes in open-set learning~\cite{open-set-survey-recenet,open-set-dis-3,open-set-dis-4}. 
The evaluation data is $D_{eval} = \{(x'_1, y'_1), (x'_2, y'_2), ..., (x'_m, y'_m)\}$, where $y'_i \in (C_B \cup \{u\})$. $u$ is a single class that represents the `unknown' class. Open-set learning tasks do not require further classifying the unknown classes.

\noindent$\bullet$
\textbf{Open World Learning.}
Open world learning addresses real-world environments' dynamic and ever-evolving nature by recognizing and learning new categories incrementally over time, without the need for complete system retraining~\cite{open-world-recognition,open-world-detection}. This process includes classifying known objects, identifying unknowns, labeling the unknowns by humans, and incrementally learning new categories as they are labeled and added to the system. Suppose an open world learning process contains $T$ learning steps. In the $t$-th learning step, where $t \in \{1,2,...,T\}$, the training data is $D_{train}^t = \{(x_1^t, y_1^t), (x_2^t, y_2^t), ..., (x_{n_t}^t, y_{n_t}^t)\}$, where $y_{i}^t$ is in the labels annotated in the $t$-th step $C^t_{label}$, and $x_{i}^t$ is the input data corresponding to the newly added labels. The evaluation data is $D_{eval}^t = \{(x'_{1}, y'^{t}_{1}), (x'_{2}, y'^{t}_{2}), ..., (x'_{m}, y'^{t}_{m})\}$, where $y'^{t}_{i} \in (C^t_{B} \cup \{u\})$. $u$ is a single class that represents the `unknown' class. $C^t_{B}$ is the known class label set for timestamp $t$. It is an accumulation of labeled classes in the current and previous steps. $C^t_{B} = \bigcup_{k=0}^{t} C^k_{label}$. Note that evaluation input $x'_i$ keeps the same across learning steps.

\noindent$\bullet$
\textbf{Out-of-Distribution Detection.}
Out-of-distribution (OOD) detection focuses on the ability to detect data that is different in some way from the data used during training~\cite{yang2021oodsurvey}. The training data is $D_{train} = \{(x_1, y_1), (x_2, y_2), ..., (x_n, y_n)\}$ and is supposed to be sampled from a probability distribution $P_{train}(X,Y)$. During testing, the model encounters data $x'_i$ that are sampled from a different distribution $P_{OOD}(X')$, which is not represented in the training dataset. The goal is to identify or appropriately handle these OOD samples. Metrics for OOD detection often involve measuring the model’s certainty or confidence in its predictions, using scores like softmax probability, and checking whether it correctly identifies OOD samples~\cite{yang2021oodsurvey}.

\noindent$\bullet$
\textbf{Zero-Shot Learning.} Zero-shot learning aims to recognize objects or concepts not seen during training. The training data is $D_{train} = \{(x_1, y_1), (x_2, y_2), ..., (x_n, y_n)\}$, where y is from base classes $C_B$, which is usually called seen classes in zero-shot learning~\cite{zeroshotinst,zeroshotobjectdetection1}. The evaluation data is $D_{eval} = \{(x'_1, y'_1), (x'_2, y'_2), ..., (x'_m, y'_m)\}$, where $y'_i \in C_N$. $C_N$ is novel classes or unseen classes. Zero-shot learning tasks require models to make clear classifications among the new unseen classes.

We briefly compare these concepts in Fig.~\ref{fig:bg-task-diff}, including open vocabulary, open-set, open world, OOD, and zero-shot.

\subsection{History and Roadmap}

Before introducing the open vocabulary setting, reviewing the progress of open vocabulary learning is necessary. In Fig.~\ref{fig:timeline}, we summarize the timeline of open vocabulary learning. Localization and classification of arbitrary objects in the wild have been challenging problems due to the limitations of existing datasets. The concept of open vocabulary in scene understanding comes from the work~\cite{zhao2017open}, where the authors build a joint image pixel and word concept embedding framework. The concept is hierarchically divided. Then, multi-modal pre-training was well studied with the rise of BERT~\cite{devlin2018bert} in NLP. Motivated by the process of vision language pre-training, OVR-CNN~\cite{zareian2021opendet_ovrcnn} proposed the concept of open vocabulary object detection, where the caption data are used for connecting novel classes semantics and visual region. Later on, CLIP was presented and open-sourced. After that, VilD~\cite{gu2021open_vild} is the first work that uses the knowledge of CLIP to build open vocabulary object detection.
Meanwhile, LSeg~\cite{Language-driven-semantic-segmentation} first explored the CLIP knowledge of language-driven segmentation tasks. After these works, recently, there have been more and more works on improving the performance of open vocabulary detectors or building new benchmarks for various settings. SAM~\cite{kirillov2023segment} is proposed to build the segmentation foundation model, which is trained by billion-level masks. Combined with CLIP, SAM can also achieve good zero-shot segmentation without fine-tuning. Recently, with the rapid process of large language model (LLM)~\cite{touvron2023llama}, open vocabulary learning has become a more promising direction since more language knowledge can be embedded in multi-modal architecture. As shown in Fig.~\ref{fig:statistic}(a), the number of research works in open vocabulary learning has increased significantly since 2021. We also summarize the statistics of different directions in Fig.~\ref{fig:statistic}(b) and (c). The details of these directions can be found in Sec.~\ref{sec:method_survey}.

\subsection{Tasks, Datasets, and Metrics}

\noindent$\bullet$
\textbf{Tasks.} Open vocabulary learning has included a wide range of computer vision tasks, including object detection~\cite{zareian2021opendet_ovrcnn, VLDet, feng2022promptdet}, segmentation~\cite{OpenSeg, CGG, panoptic-MaskCLIP}, video understanding, and 3D scene understanding. The center goal of these tasks is similar, recognizing the novel classes with the aid of large vocabulary knowledge for their corresponding tasks. In this survey, we mainly focus on the methods of scene understanding tasks, including object detection, instance segmentation, semantic segmentation, and object tracking. Nonetheless, we also consider other closely related tasks, such as open vocabulary attribution prediction, video classification, and point cloud classification.

\noindent$\bullet$
\textbf{Datasets.} For object detection, the common datasets are COCO~\cite{COCO_dataset} and LVIS~\cite{gupta2019lvis}.
Recently, a more challenging dataset, v3Det~\cite{wang2023v3det} with more than 10,000 categories, is proposed.
For image segmentation, the most commonly used datasets are COCO~\cite{COCO_dataset}, ADE20k~\cite{ADE20K}, PASCAL-VOC 2012~\cite{pascal-voc-dataset}, PASCAL-Context~\cite{pascal-context-dataset}, and Cityscapes~\cite{cordts2016cityscapes}. For video segmentation and tracking, the frequently used datasets are VSPW~\cite{miao2021vspw}, Youtube-VIS~\cite{vis_dataset}, LV-VIS~\cite{wang2023towards}, MOSE~\cite{MOSE}, and TAO~\cite{dave2020tao}.

\noindent$\bullet$
\textbf{Metrics.} For detection tasks, the commonly used metrics are mean average precision (mAP) and mean average recall (mAP) for both base and novel classes. Among segmentation tasks, the commonly used metrics are mean intersection over union (mIoU) for semantic segmentation, mask-based mAP for instance segmentation, and panoptic quality (PQ) for panoptic segmentation.

\subsection{Related Research Domains}

\noindent$\bullet$
\textbf{Open-Set Recognition.}
The concept of Open-Set Recognition (OSR) addresses the challenge of identifying unknown classes during classification tasks. In traditional classification systems, models are trained to recognize a finite set of known classes, but in real-world scenarios, they may encounter data that doesn't belong to these predefined categories. During testing, OSR aims to classify known classes seen during the training and reject unknown classes that are unseen~\cite{open-set-survey-recenet}. The main methods used for OSR can be broadly categorized into two types: discriminative models~\cite{open-set-dis-1,open-set-dis-2,open-set-dis-3,open-set-dis-4,open-set-dis-5} and generative models~\cite{open-set-gen-i-1,open-set-gen-i-2,open-set-gen-i-3,open-set-gen-i-4,open-set-gen-i-5,open-set-gen-ni-1}. The discriminative models focus on differentiating between known classes and identifying unknown classes by enhancing the boundary or margin between these classes. The generative models are either instance generation-based~\cite{open-set-gen-i-1,open-set-gen-i-2,open-set-gen-i-3,open-set-gen-i-4,open-set-gen-i-5} or non-instance generation-based~\cite{open-set-gen-ni-1}. They emphasize generating new instances or features to improve the ability of the system to recognize new, unseen classes during testing. Each category employs specific techniques and approaches to address the challenges inherent in open-set recognition, which involves dealing with classes not seen during the training phase. ~\cite{open-set-extend} extends the OSR task to require further identifying novel classes.

\noindent$\bullet$
\textbf{Open World Learning.}
Open world learning involves identifying and labeling new, unknown categories (novel unknowns)~\cite{open-world-recognition}. This process includes incrementally learning new categories as they are labeled and added to the system. The goal is to create a system that remains robust to unknown categories and adapts continually to include new information, balancing the risks associated with open spaces in the learning model. \cite{open-world-recognition} first propose open word recognition. Recent works also explore the open world object detection~\cite{openworld-detection, open-world-detection-2} task.

\noindent$\bullet$
\textbf{Out-of-Distribution Detection.}
Out-of-distribution (OOD) detection methods can be structured into several categories~\cite{yang2021oodsurvey}. Classification-Based Methods~\cite{ood-cls-output-1,ood-cls-output-2,ood-cls-labelspace-1,ood-cls-gen-1}: These include output-based methods~\cite{ood-cls-output-1,ood-cls-output-2}, label space redesign~\cite{ood-cls-labelspace-1}, and OOD data generation techniques~\cite{ood-cls-gen-1}. Density-Based Methods~\cite{ood-density-1,ood-density-2,ood-density-3}: This category involves methods that detect OOD by modeling data density. Distance-Based Methods~\cite{ood-distance-1,ood-distance-2}: These methods use distance metrics, typically in the feature space, to identify OOD instances. Reconstruction-Based Methods~\cite{ood-re-1,ood-re-2,ood-re-3}: This approach achieves OOD detection that features reconstruction capabilities.

\noindent
$\bullet$ \textbf{Zero-Shot Detection and Segmentation:} This task aims to segment classes that have not been encountered during training. Two streams of work have emerged: discriminative methods~\cite{spnet,PMOSR,D2Zero,PAP3D,baek2021exploiting} and generative methods~\cite{PADing,ZS3Net,CaGNet,shen2021conterfactual}. Representative works in this field include SPNet~\cite{spnet} and ZS3Net~\cite{ZS3Net}.
SPNet~\cite{spnet} maps each pixel to a semantic word embedding space and projects pixel features onto class probabilities using a fixed semantic word embedding~\cite{fasttext,word2vec} projection matrix. On the other hand, ZS3Net~\cite{ZS3Net} first trains a generative model to produce pixel-wise features for unseen classes based on word embeddings. With these synthetic features, the model can be trained in a supervised manner. Both of these works treat zero-shot detection and segmentation as a pixel-level zero-shot classification problem. However, this formulation is not robust for zero-shot learning, as text embeddings are typically used to describe objects/segments rather than individual pixels. Subsequent works~\cite{CSRL,CaGNet,CaGNetv2,PADing,shen2021conterfactual,cap2seg} follow this formulation to address different challenges in zero-shot learning.
In a weaker assumption where unlabeled pixels from unseen classes are available in the training images, self-training~\cite{ZS3Net} is commonly employed. Despite promising results, self-training often requires model retraining whenever a new class appears. ZSI~\cite{zeroshotinst} also employs region-level classification for bounding boxes but focuses on instance segmentation rather than semantic segmentation. More recently, PADing~\cite{PADing} proposes a unified framework to tackle zero-shot semantic segmentation, zero-shot instance segmentation, and zero-shot panoptic segmentation.

Most approaches in open vocabulary learning are based on zero-shot learning settings, such as replacing the fixed classifier with language embeddings. 
However, these methods struggle to generalize well to novel classes due to the absence of novel class knowledge. 
Consequently, their performance is limited, and they are not practical for real-world applications.

\noindent$\bullet$
\textbf{Long-tail Object Detection and Instance Segmentation.} This task addresses the challenge of class imbalance in instance segmentation. 
Many approaches tackle this issue through techniques such as data re-sampling~\cite{gupta2019lvis,liu2020deep}, loss re-weighting~\cite{ren2020balanced}, and decoupled training~\cite{li2020overcoming}. 
Specifically, some studies~\cite{liu2020deep} employ image-level re-sampling. However, these methods tend to introduce bias in instance co-occurrence. 
To address this issue, other works~\cite{hu2020learning} focus on more refined re-sampling techniques at the instance or feature level. Regarding loss re-weighting, most studies~\cite{ren2020balanced} rebalance the ratio of positive and negative samples during training. 
Additionally, decoupled training methods~\cite{li2020overcoming,wang2020devil} introduce different calibration frameworks to enhance classification results. 
Resolving long-tail object detection can lead to improved accuracy of rare classes. However, these methods currently cannot be applied to the detection of novel classes.

\noindent
$\bullet$ \textbf{Few-Shot Detection and Segmentation:} Few-shot object detection aims to expand the detection capabilities of a model using only a few labeled samples. 
Several approaches~\cite{li2021few,lee2022few,wu2021generalized,hu2023suppressing} have been proposed to advance this field. 
Notably, TFA~\cite{wang2020frustratingly} introduces a simple two-phase fine-tuning method, while DeFRCN~\cite{qiao2021defrcn} separates the training of RPN features and RoI classification. 
Moreover, SRR-FSD~\cite{zhu2021semantic} combines multi-modal inputs while LVC~\cite{kaul2022label} proposes a pipeline to generate additional examples for novel object detection and train a more robust detector. 
Few-shot segmentation comprises Few-Shot Semantic Segmentation (FSSS) and Few-Shot Instance Segmentation (FSIS). 
FSSS involves performing pixel-level classification on query images. Previous approaches~\cite{ SRPNet,rakelly2018conditional} typically build category prototypes from support images and segment the query image by computing the similarity distance between each prototype and query features. 
On the other hand, FSIS aims to detect and segment objects with only a few examples. FSIS methods can be categorized into single-branch and dual-branch methods. 
The former~\cite{ganea2021incremental} primarily focuses on designing the classification head, while the latter~\cite{michaelis2018one, fan2020fgn, yan2019meta} introduce an additional support branch to compute class prototypes or re-weighting vectors of support images. 
This support branch assists the segmenter in identifying target category features through feature aggregation. For example, Meta R-CNN~\cite{yan2019meta} performs channel-wise multiplication on RoI features, and FGN~\cite{fan2020fgn} aggregates channel-wise features at three stages: RPN, detection head, and mask head.
However, few-shot learning still requires examples of novel classes during training, and such data may not be available.

\section{Methods: A Survey}
\label{sec:method_survey}

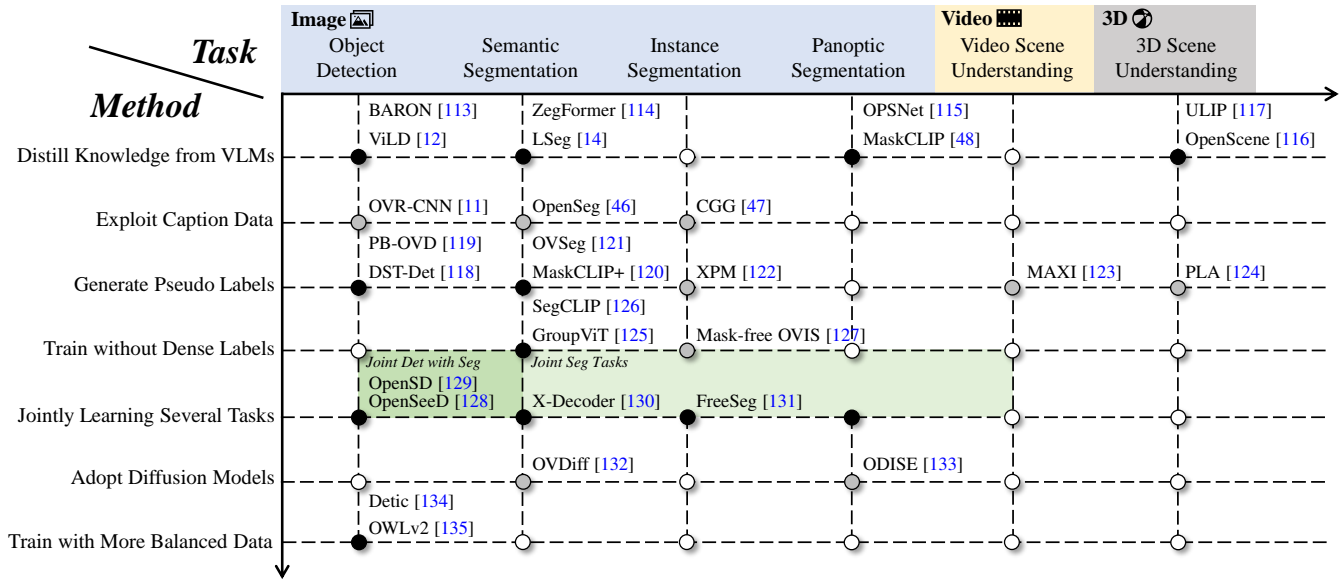
\begin{figure*}[!t]
	\centering
    \includegraphics[width=1\linewidth]{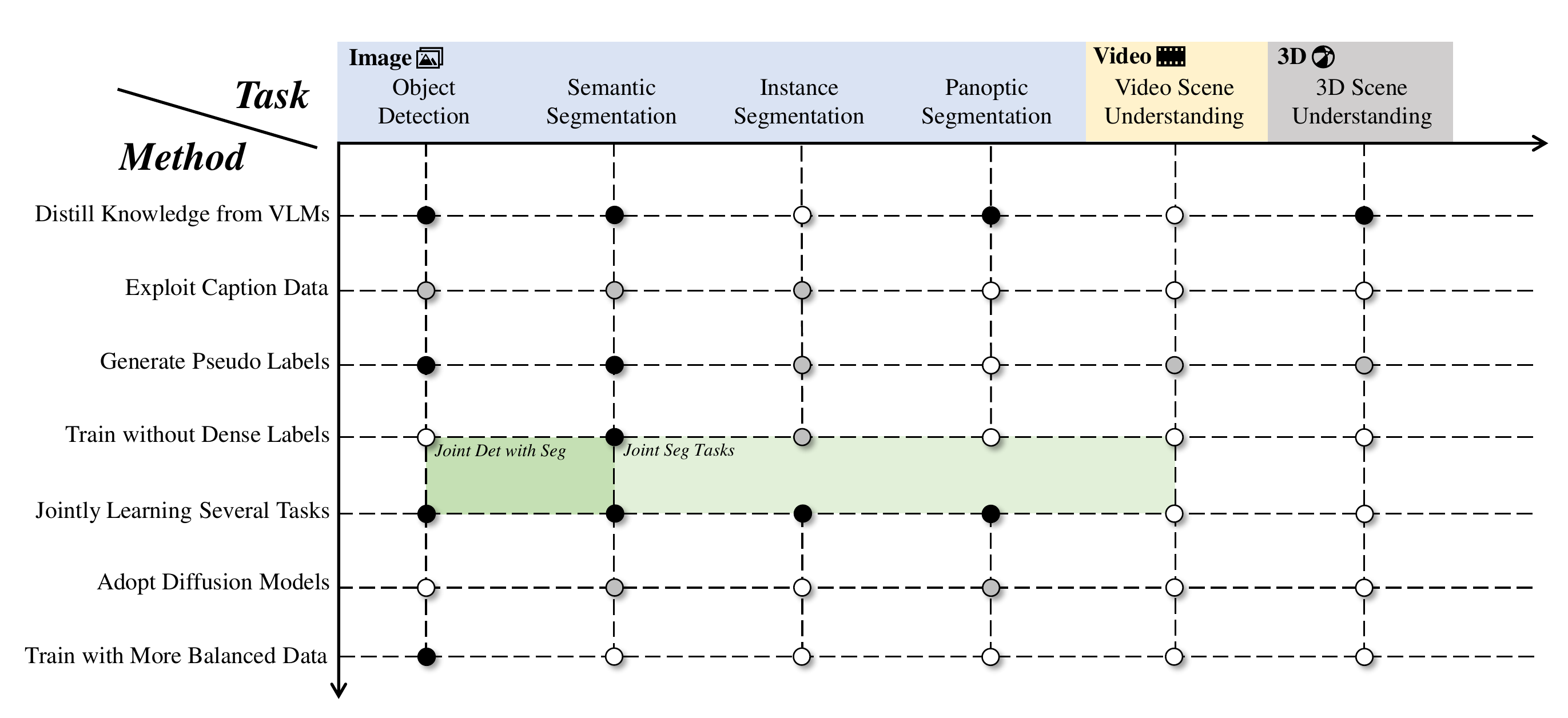}
    \begin{picture}(0,0)
        \scriptsize {
            \put(-113,182){ViLD~\cite{gu2021open_vild}}
            \put(-113,193){BARON~\cite{wu2023baron}}
            \put(-51,182){LSeg~\cite{Language-driven-semantic-segmentation}}
            \put(-51,193){ZegFormer~\cite{Decoupling-zero-shot-semantic-segmentation}}
            \put(74,182){MaskCLIP~\cite{panoptic-MaskCLIP}}
            \put(74,193){OPSNet~\cite{OPSNet}}
            \put(196, 182){OpenScene~\cite{Peng2023OpenScene}}
            \put(196, 193){ULIP~\cite{xue2023ulip}}
            \put(-113,157){OVR-CNN~\cite{zareian2021opendet_ovrcnn}}
            \put(-51,157){OpenSeg~\cite{OpenSeg}}
            \put(11,157){CGG~\cite{CGG}}
            \put(-113,132){DST-Det~\cite{Xu2023DSTDetSD}}
            \put(-113,143){PB-OVD~\cite{Gao2021pbovd}}
            \put(-51,132){MaskCLIP+~\cite{Denseclip}}
            \put(-51,143){OVSeg~\cite{Mask-adapted-clip}}
            \put(11,132){XPM~\cite{XPM}}
            \put(136, 132){MAXI~\cite{lin2023match}}
            \put(196, 132){PLA~\cite{ding2023language}}
            \put(-51,108){GroupViT~\cite{Groupvit}}
            \put(-51,119){SegCLIP~\cite{SegCLIP}}
            \put(11,108){Mask-free OVIS~\cite{mask-free-OVIS}}
            \put(-113,83){OpenSeeD~\cite{OpenSeeD}}
            \put(-113,90){OpenSD~\cite{ov-seg-unify-opensd}}
            \put(-51,83){X-Decoder~\cite{X-decoder}}
            \put(11,83){FreeSeg~\cite{freeseg}}
            \put(-51,59){OVDiff~\cite{diffusion_sem}}
            \put(74,59){ODISE~\cite{ODISE}}
            \put(-113,45){Detic~\cite{zhou2022detecting}}
            \put(-113,35){OWLv2~\cite{owlv2}}
        }
    \end{picture}
    \vspace{-20pt}
    \caption{Open vocabulary learning methods, organized by their tasks and approach types. We list several representative works here.}
	\label{fig:method-taxonomy}
\end{figure*}

\begin{figure}[!t]
	\centering
	\includegraphics[width=1\linewidth]{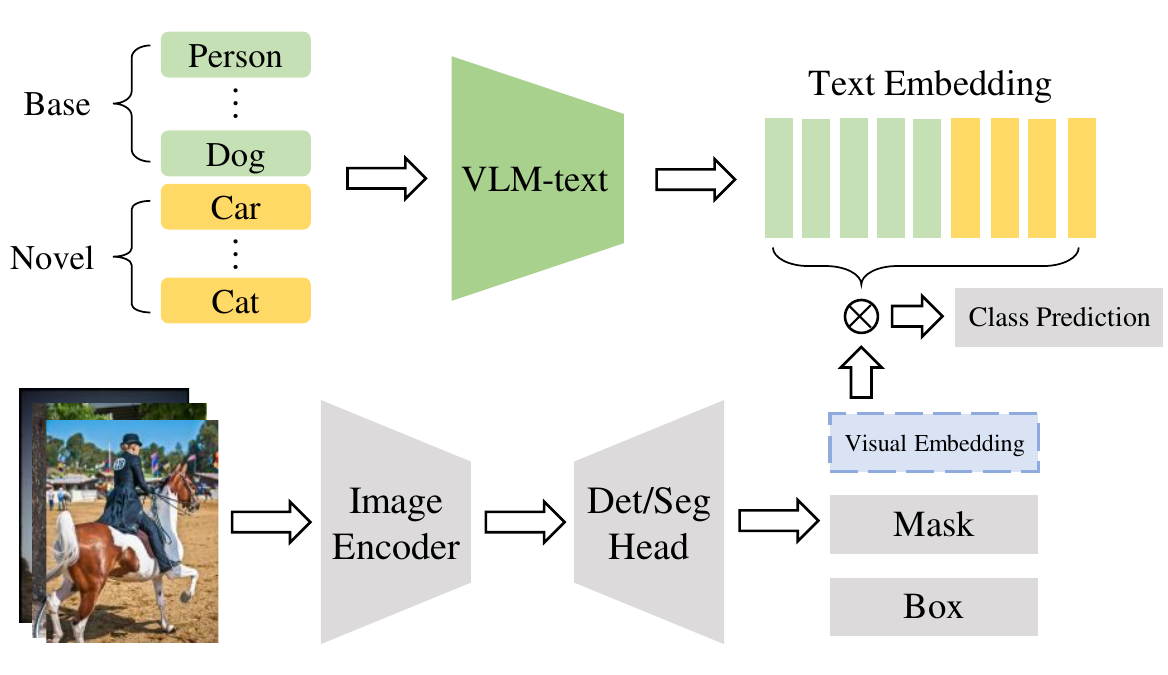}
        \caption{An illustration of a common architecture in open vocabulary object detection and segmentation. The vision model predicts a class embedding for each box/mask. The embeddings are compared to a set of class embeddings generated by a VLM text model like CLIP or ALIGN, using dot products. The class with the highest score is chosen as the predicted label for the object. Note that while humans define the set of possible object classes, the system only has access to a limited set of ``base" classes during training.}
	\label{fig:common-architecture}
\end{figure}

\noindent
\textbf{Overview.} In this section, we first review preliminary knowledge and vision language modeling by extending close-set detectors into open vocabulary detectors via VLMs in Sec.~\ref{sec:pre_knowledges} and Sec.~\ref{sec:vlm}.
Then we sequentially survey the six subsidiary tasks, including object detection~ (Sec.~\ref{sec:open_voc_object_det}), segmentation (Sec.~\ref{sec:open_voc_seg}), video understanding (Sec.~\ref{sec:open_ov_videos}), 3D scene understanding (Sec.~\ref{sec:open_voc_3d}), and closely related tasks (Sec.~\ref{sec:closely_related_tasks}). Note that we only record and compare the most representative works. We list numerous works in Tab.~\ref{tab:det_seg_method_summary} and Tab.~\ref{tab:video_3d_summary}. Moreover, since there are several similar tasks (Sec.~\ref{sec:closely_related_tasks}), we also survey and compare other related tasks, including class agnostic detection and segmentation, open world object detection, and open-set panoptic segmentation.

\noindent
\textbf{Method taxonomy and relation of each subsection.} We summarize the most commonly used methods among these different directions in Fig.~\ref{fig:method-taxonomy}. However, we cannot include all methods due to the wide range of tasks. In each subsection, we summarize different methods according to different problems, segmentation, detection, video understanding, and closely related topics. We cluster methods according to common practice, such as knowledge distillation and region text pre-training. We start with object detection since it was first proposed. For segmentation tasks, we ignore the specific setting in Sec.~\ref{sec:open_voc_seg}. We argue that despite these two directions, they may share similar ideas. However, most meta-architectures and tasks are different. Thus, we still review each direction individually. For video, 3D understanding, and other closely related topics, due to different task definitions and extra input cues, such as temporal information and multi-view inputs, we also survey two directions individually. 

\noindent
\textbf{Motivation of survey organization.} We argue that using the unified symbolic system to summarize all methods is hard. There are several reasons. \textbf{(1)}, The task definitions and settings are different. In addition to different input and output formats, taking the open vocabulary segmentation as an example, there are at least three different settings~\cite{Groupvit,CGG,OpenSeg} for open vocabulary segmentation. Thus, the design principles are different. For example, OpenSeg~\cite{OpenSeg} uses supervised pretraining and contrastive losses while GroupViT~\cite{Groupvit} adopt unsupervised setting without mask annotations. \textbf{(2)}, Different methods use different detectors or baseline models. Several works~\cite{zareian2021opendet_ovrcnn,zhou2022detecting} use R-CNN framework while several methods~\cite{CGG,freeseg} adopt query-based approaches. Several methods~\cite{dst_ov_det,Kuo2022FVLMOO} adopt the RPN's features to develop their methods, while query-based methods cannot perform these designs. \textbf{(3)}, Different methods may also use different datasets during training. Several methods~\cite{zhou2022detecting,MosaicFusion,wu2023diffumask} mainly explore the dataset effect on open-vocabulary detection or segmentation. Since the supervision signals are different, it will hard to put all methods in one unified symbolic system. In summary, our survey focus on more comprehensive review to extract the common features of various domains in open vocabulary learning.

\subsection{Preliminary}
\label{sec:pre_knowledges}

\noindent 
\textbf{Pixel-based Object Detection and Segmentation.} In general, it can be divided into two aspects: \textit{semantic-level} tasks and \textit{instance-level} tasks, where the difference lies in whether to distinguish each instance. For the former, we take semantic segmentation as an example. It was typically approached as a dense pixel classification problem, as initially proposed by FCN~\cite{ding2021interaction}. Then, the following works are all based on the FCN framework. These methods can be divided into the following categories, including better encoder-decoder frameworks~\cite{yu2018learning,ding2020semantic}, larger kernels~\cite{peng2017large}, multiscale pooling~\cite{zhao2017pyramid,deeplabv3}, multiscale feature fusion~\cite{ding2018context,li2020gated,sfnet,Li2022SFNetFA}, non-local modeling~\cite{wang2018nonlocal}, and better boundary delineation~\cite{kirillov2020pointrend,BoundaryAware,li2020improving,he2021boundarysqueeze}. After the transformer was proposed, with the goal of global context modeling, several works propose the variants of self-attention operators to replace the CNN prediction heads~\cite{DAnet,hu2018relation}. 

For the latter, we take object detection and instance segmentation for illustration. Object detection aims to detect each instance box and classify each instance. It mainly has two different categories: two-stage approaches and one-stage approaches. Two-stage approaches~\cite{ren2015faster,maskrcnn} rely on an extra region proposal network (RPN) to recall foreground objects at the first stage. Then, the proposals (Region of Interests, RoI) with high scores are sent to the second-stage detection heads for further refinement. One-stage approaches~\cite{tian2019fcos,focal_loss} directly output each box and label in a per-pixel manner. In particular, with the help of focal loss and feature pyramid networks, one-stage approaches can surpass the two-stage methods in several datasets. Both approaches need anchors for the regression of the bounding box. The anchors are the pixel locations where the objects are possibly emerging. Since two-stage approaches explicitly explore the foreground objects, where they have better recall than single-stage approaches. Two-stage approaches are more commonly used in open vocabulary object detection tasks.

Instance segmentation segmenting each object goes beyond object detection. 
Most instance segmentation approaches focus on how to represent instance masks beyond object detection, which can be divided into two categories: top-down approaches~\cite{maskrcnn,tian2020conditional} and bottom-up approaches~\cite{neven2019instanceSeg,de2017semanticInstanceLoss}. 
The former extends the object detector with an extra mask head. The designs of mask heads are various, including FCN heads~\cite{maskrcnn,htc}, diverse mask encodings~\cite{zhang2020MEInst}, and dynamic kernels~\cite{tian2020conditional,bolya2019yolact}. 
The latter performs instance clustering from semantic segmentation maps to form instance masks.
The performance of top-down approaches is closely related to the choice of detectors~\cite{qiao2021detectors}, while bottom-up approaches depend on both semantic segmentation results and clustering methods~\cite{cheng2020panoptic}. 
Besides, several approaches~\cite{wang2020solov2} use gird representation to learn instance masks directly. 

\noindent
\textbf{Query-based Object Detection and Segmentation.} With the rise of vision transformers~\cite{VIT, detr}, recent works mainly use transformer-based approaches in segmentation, detection, and video understanding. Compared with previous pixel-based approaches, transformer-based approaches have more advantages in cases of flexibility, simplicity, and uniformity~\cite{detr,zhou2022transvod,ding2022vlt,GRES,panopticpartformer,li2023panopticpartformer++,yuan2021polyphonicformer,xu2022fashionformer,xu2022multi,MOSE,tubelink,hu2023dacdetr}.
One representative work is detection transformer (DETR)~\cite{detr}. It contains a CNN backbone, a standard transformer encoder, and a standard transformer decoder. It also introduces the concepts of object query to replace the anchor design in pixel-based approaches.
Object query is usually combined with bipartite matching~\cite{kuhn1955hungarian} during training, uniquely assigning predictions with ground truth. 
This means each object query builds the one-to-one matching during training. Such matching is based on the matching cost between ground truth and predictions. The matching cost is defined as the distance between prediction and ground truth, including labels, boxes, and masks. 
By minimizing the cost with the Hungarian algorithm~\cite{kuhn1955hungarian}, each object query is assigned by its corresponding ground truth. 
For object detection, each object query is trained with classification and box regression loss~\cite{ren2015faster}. 
For instance-wised segmentation, each object query is trained with classification loss and segmentation loss. The output masks are obtained via the inner product between object query and decoder features. 
Recently, mask transformers~\cite{wang2020maxDeeplab,zhang2021knet,cheng2021maskformer} further removed the box head for segmentation tasks. Max-Deeplab~\cite{wang2020maxDeeplab} is the first to remove the box head and design a pure-mask-based segmenter. It combines a CNN-transformer hybrid encoder~\cite{axialDeeplab} and a query-based decoder as an extra path. Max-Deeplab still needs many auxiliary loss functions. 
Later, K-Net~\cite{zhang2021knet} uses mask pooling to group the mask features and designs a dynamic convolution to update the corresponding query. %
Meanwhile, MaskFormer~\cite{cheng2021maskformer} extends the original DETR by removing the box head and transferring the object query into the mask query via MLPs.
It proves simple mask classification can work well enough for all three segmentation tasks.
Then, Mask2Former~\cite{cheng2021mask2former} proposes masked cross-attention and replaces the cross-attention in MaskFormer. 
Masked cross-attention forces the object query only attends to the object area, guided by the mask outputs from previous stages. 
Mask2Former also adopts a stronger Deformable FPN backbone~\cite{zhu2020deformabledetr}, stronger data augmentation~\cite{detectron2}, and multiscale mask decoding. 
In summary, query-based approaches are stronger and simpler. They cannot directly detect novel classes but are widely used as base detectors and segmenters in open vocabulary settings.

\subsection{Vision Language Modeling}
\label{sec:vlm}

\noindent
\textbf{Large Scale Visual Language Pre-training.} Better visual language pre-training can lead to a better understanding of semantics on given visual inputs~\cite{ding2022vlt,VLT_iccv2021,GRES}. Previous works~\cite{li2019visualbert,li2020unicoder,lu2019vilbert} focus on cross-modality research via learning the visual and sentence-dense connection between different modalities. 
Some works~\cite{tan2019lxmert,lu2019vilbert} utilize two-stream neural networks based on the vision transformer model, while several works~\cite{qi2020imagebert,huang2020pixelbert} adopt the single-stream neural network, where the text embeddings are frozen.
The two-stream neural networks process visual and language information and fuse them afterward by another transformer module.
Recently, most approaches~\cite{CLIP,jia2021scaling_align} adopt pure transformer-based visual language pre-training. 
Both CLIP~\cite{CLIP} and Align~\cite{jia2021scaling_align} are concurrent works that explore extremely large-scale pre-training on image text pairs.
In particular, with such large-scale training, CLIP demonstrates that the simple pre-training task of predicting which caption goes with which image can already lead to stronger generalizable models.
Several following works~\cite{li2023scaling,mu2022slip,sun2023evaclip} aim to improve the CLIP training via mask image modeling, scaling up the training data and model size.
These VLMs are the foundation of open vocabulary learning in different tasks, which means that the open vocabulary approaches aim to distill or utilize knowledge of VLMs into their corresponding tasks.

Moreover, in addition to achieving better zero-shot recognition, several works focus on designing better vision language models for language-related tasks, including visual question answering (VQA). 
Several works~\cite{ALBEF, yu2022coca, yan2022video, li2022blip, li2023blip2} explore how to better align caption loss and contrastive loss during the image-text pertaining.
In particular, CoCa~\cite{yu2022coca} adopts cross attention to connect caption generation part and contrastive learning. 
Recently, BLIP-2~\cite{li2023blip2} bootstraps vision-language pre-training from off-the-shelf frozen pre-trained image encoders and frozen large language models via a lightweight Querying Transformer.

\noindent$\bullet$
\textbf{Visual Grounding Tasks.} Visual grounding tasks aim to localize the specific objects according to given text descriptions. 
Referring segmentation tasks~\cite{GRES,refcoco,robustRefSeg,MeViS} segment the specific object.
Referring expression comprehension~\cite{TransVG,li2023iterative, wang2023ov} localizes the bounding boxes of given object texts. 
Recent research is mainly based on two-stream networks: one for visual encoder~\cite{resnet,VIT} and the other for text encoder~\cite{devlin2018bert}.
These works focus on how to better match text features and visual features via different architectures. 
For referring segmentation, previous works~\cite{CMPC, MCN, VLT_iccv2021} adopt ``decoder-fusion'', where they design a separate decoder at the end of the two-stream networks.
Recent works~\cite{EFN, yang2022lavt} explore the ``encoder-fusion'' to directly fuse the text features into a visual backbone before mask prediction.
For referring expression comprehension, similar approaches~\cite{TransVG, Ref-NMS} are adopted such as encoder-fusion.
Recently, several works~\cite{mdetr,liu2023groundingdino} aim to unify visual grounding and object detection.
In particular, Grounding DINO~\cite{liu2023groundingdino} unifies detection and grounding in one framework. It uses stronger detectors with multiple dataset pre-training, which achieves strong results for many downstream tasks. 
Compared with open vocabulary learning, visual grounding tasks require visual text matching with specific text descriptions. 
Open vocabulary learning tasks require the model to automatically detect, segment and recognize new objects without the given text information, such as class names, which is more challenging.

\noindent
\textbf{Turning Close-set Detector and Segmenter Into Open Vocabulary Setting.} A common way towards the open vocabulary setting is to replace the fixed classifier weights with the text embeddings from a VLM model. In Fig.~\ref{fig:common-architecture}, we present a meta-architecture. In particular, the vision model generates a visual embedding for each box/mask proposal and computes similarity scores by computing a dot product with the text embeddings from both base and novel classes. The classification scores are computed as follows:
\begin{equation}
   p_{ij} = \frac{\exp{\langle e^v_i, e^t_j \rangle}}{1 + \sum\limits^{|C_B \cup C_N|}_{j'=1} \exp{\langle e^v_i, e^t_{j'} \rangle}}, 
\end{equation}
where $e^v_i$ is the $i$-th vision embedding and $e^t_j$ is the $j$-th text embedding, and $p_{ij}$ is the prediction score of the $i$-th vision proposal predicted to the $j$-th class. The $\langle ., . \rangle$ represents the dot product operation. The denominator is added with one because the background text embedding is set to all 0 or make it learnable. $C_B$ and $C_N$ are base and novel classes, respectively. Finally, the class with the highest prediction score is chosen as the prediction. In practice, the vision embeddings are trained with base annotations to fit the text embeddings. Therefore, the model combines the knowledge of VLM and the learned visual features, and the detector/segmenter can detect/segment novel classes via semantically related text embeddings.

\begin{table*}[ht]
    \centering
    \tiny
    \caption{Representative works summarization and comparison in Sec.~\ref{sec:open_voc_object_det} and Sec.~\ref{sec:open_voc_seg}. We list the training data, detectors or segmenters, VLMs, and highlighted features for comparison.}\vspace{-3mm}
    \setlength{\tabcolsep}{1.2pt}
    \scalebox{0.98}{
\begin{tabular}{p{0.10\textwidth}p{0.05\textwidth}p{0.10\textwidth}p{0.20\textwidth}p{0.10\textwidth}p{0.15\textwidth}p{0.30\textwidth}}
    \toprule
    \belowrulesepcolor{gray!30!}
\rowcolor{gray!30!} Method & Task & Text Training Data & Vision Training Annotations & Text Model & Vision Model & \ \ ~~~~~~~~~~~~~~~~~~~~~~~~~~~~~~~~~~~~Highlight \\ \aboverulesepcolor{gray!30!} \midrule
\belowrulesepcolor{orange!15!}
\rowcolor{orange!15!}\multicolumn{7}{c}{\textbf{Open Vocabulary Detection (Sec.~\ref{sec:open_voc_object_det})}} \\ \aboverulesepcolor{orange!15!} \midrule
OVR-CNN~\cite{zareian2021opendet_ovrcnn} & OVOD & Captions & Bounding Boxes (Base) & BERT & Faster R-CNN & The first method that proposes OVOD and adopts grounded caption pre-training.  \\ 
ViLD~\cite{gu2021open_vild} & OVOD & None & Bounding Boxes (Base)  & CLIP-text & CLIP-vision + Faster R-CNN & The first method that distills knowledge from the pre-trained CLIP model. \\
VL-PLM~\cite{zhao2022VLPLM} & OVOD & None & Bounding Boxes (Base)  & CLIP-text & CLIP-vision + Faster R-CNN & Generate pseudo-labels for novel classes using pre-trained OVOD. \\
RegionCLIP~\cite{Zhong2021RegionCLIPRL} & OVOD & Captions & Bounding Boxes (Base + Pseudo Novel) & CLIP-text & CLIP-vision + Faster R-CNN & Creates region-text pairs as pseudo labels using CLIP and pre-train the detector in the first stage. \\
Detic~\cite{zhou2022detecting} & OVOD & ImageNet & Bounding Boxes (Base + Pseudo Novel) & CLIP-text & Centernet2 & Propose a weakly supervised approach that is training rare classes with image-level annotations.  \\
DetPro~\cite{Du2022LearningTP} & OVOD & ImageNet & Bounding Boxes (Base + Pseudo Novel) & CLIP-text & Centernet2 & learn continuous prompt representations for open vocabulary object detection based on the pre-trained vision-language model. \\
OV-DETR~\cite{zang2022open} & OVOD & None & Bounding Boxes (Base) & CLIP-text & CLIP-vision + DETR & Introduces a conditional binary matching mechanism to let DETR model generalize to queries from unseen classes. \\
CORA~\cite{wu2023cora} & OVOD & None & Bounding Boxes (Base) & CLIP-text &  CLIP-vision + DETR & Propose Anchor Pre-Matching strategy to reduce both training and inference time for conditional binary matching. \\
VLDet~\cite{VLDet} & OVOD & Captions & Bounding Boxes (Base) & CLIP-text & Faster R-CNN & Aligns image regions with words in captions by a set matching method. \\
F-VLM~\cite{Kuo2022FVLMOO} & OVOD & None & Bounding Boxes (Base) & CLIP-text & CLIP-vision + Mask R-CNN & Train the detector with frozen VLMs and combine scores of joint detection and VLMs.  \\
BARON~\cite{wu2023baron} & OVOD & None & Bounding Boxes (Base) & CLIP-text & Faster R-CNN & Aligning Bag of Regions for Open Vocabulary Object Detection.  \\
OWLv2~\cite{owlv2} & OVOD & None & Bounding Boxes (Base + Pseudo Novel) & CLIP-text & ViT + detection head & Generate pseudo-labels from WebLI dataset and train the detector with the generated datasets.  \\
MaMMUT~\cite{kuo2023mammut} & OVOD & Captions & Bounding Boxes (Base + Pseudo Novel) & CLIP-text & ViT + detection head & Joint pre-train with multi-modal tasks to benefit the novel object detection.  \\

\midrule\belowrulesepcolor{orange!15!}
\rowcolor{orange!15!}\multicolumn{7}{c}{\textbf{Open Vocabulary Segmentation (Sec.~\ref{sec:open_voc_seg})}} \\ \aboverulesepcolor{orange!15!} \midrule
LSeg~\cite{Language-driven-semantic-segmentation} & OVSS & None & Segmentation Masks (Base) & CLIP-text & ViT & Aligns text embeddings from the VLM model with pixel features. \\
ZegFormer~\cite{Decoupling-zero-shot-semantic-segmentation} & OVSS & None & Segmentation Masks (Base) & CLIP-text & CLIP-vision + Query-based Transformer Decoder & Decouple segmentation and classification by generating class-agnostic segment masks then classify each mask. \\
PADing~\cite{PADing} & OVSS & None & Segmentation Masks (Base) & CLIP-text & Mask2Former & Introduce a generative model to synthesize features for unseen categories and achieve universal open-vocabulary segmentation. \\
OpenSeg~\cite{OpenSeg} & OVSS & Captions & Segmentation Masks (Base) & ALIGN & EfficientNet-B7 & Performs region-word grounding loss between mask features and word features. \\
MaskCLIP+~\cite{Denseclip} & OVSS & None & Segmentation Masks (Pseudo All) & CLIP-text & CLIP-vision + DeepLabv2 & Modifies CLIP so that it can output per-pixel feature maps. \\
OVSeg~\cite{Mask-adapted-clip} & OVSS & Captions & Segmentation Masks (Base + Pseudo All) & CLIP-text & CLIP-vision + MaskFormer & Uses CLIP to match the proposed image regions with nouns in the captions to generate pseudo labels. \\
GroupVit~\cite{Groupvit} & OVSS & Captions & None & Transformer Encoder & ViT-S & Learns semantic segmentation only with caption data. \\
Vil-Seg~\cite{Vil-Seg} & OVSS & Captions & None & ViT-B & ViT-B & Learns semantic segmentation without pixel-level annotations using contrastive loss and clustering loss. \\
CGG~\cite{CGG} & OVIS & Captions & Segmentation Masks (Base) & BERT & Mask2Former & Fully exploits caption data using caption grounding and generation. \\

MaskCLIP~\cite{panoptic-MaskCLIP} & OVPS & None & Segmentation Masks (Base) & CLIP-text & CLIP-vision + Mask2Former & Proposes Relative Mask Attention (RMA) modules to adapt cropped images to the pre-trained CLIP model. \\
ODISE~\cite{ODISE} & OVPS & Captions & Segmentation Masks (Base) & CLIP-text & Stable Diffusion & Exploits the vision-language alignment learned by denoising diffusion models. \\
OVDiff~\cite{diffusion_sem} & OVSS & None & None & CLIP-text & Stable Diffusion & Proposes a prototype-based method. Use diffusion models to produce prototypes.\\
OpenSeed~\cite{OpenSeeD} & OVIS & None & Segmentation Masks \& Boxes (Base) & UniCL & MaskDINO & Jointly learns from detection and segmentation data. \\
OVSegmentor~\cite{OVSegmentor} & OVSS & Captions & None & BERT & DINO & introduces masked entity completion and cross-image mask constituency objectives to improve training. \\

\bottomrule

\end{tabular}
}
\label{tab:det_seg_method_summary}
\vspacefigtext
\end{table*}

\begin{table*}[!t]
   \centering
    \caption{Feature summarization of current open vocabulary object detection approaches.}\vspace{-3mm}
   \scalebox{0.90}{{\begin{tabular}{c | c c }
      \toprule[0.15em]
        Method & Common Features   \\
        \midrule[0.15em]
        \rowcolor{gray!15} Knowledge Distillation & Distil the aligned visual-text knowledge of VLM into object detection. Focus on distilling loss design and targe proposal generation.   \\
        \rowcolor{orange!15} Region Text Pre-training & Use large-scale and easily available text-image pairs to pre-train a better and universal detector and finetune on the target datasets \\
        \rowcolor{red!15} Training with More Balanced Data & Adopt extra data or balanced datasets as an augmentation to train the classification head. \\ 
       \rowcolor{yellow!15} Prompting Modeling  & Adopt different prompting modeling to better transfer the VLM knowledge into detector.  \\
        \rowcolor{orange!15} Region Text Alignment & Design a better region-text alignment method to better align the region visual features (RoIs) into text features of VLM. \\
      \bottomrule
   \end{tabular}}}
   \label{tab:ovd_summary}
\end{table*}

\subsection{Open Vocabulary Object Detection}
\label{sec:open_voc_object_det}

In this section, we divide the methods into five categories: knowledge distillation, region text pre-training, training with more balanced data, prompting modeling, and region text alignment. Finally, we summarize the common features in Tab.~\ref{tab:ovd_summary}.

\noindent
\textbf{Knowledge Distillation.} These techniques aim to distill the knowledge of Vision-and-Language Models (VLMs) into close-set detectors~\cite{ren2015faster,maskrcnn,detr}. 
Since the knowledge of VLMs is much larger than close-set detectors, distilling novel classes into based classes trained detectors is a straightforward idea.
Knowledge distillation aims to distill visual knowledge directly into close-set detectors since the visual features are aligned with text features during the VLM pre-training stage.
One earlier method is ViLD~\cite{gu2021open_vild}, a two-stage detection approach that utilizes instance-level visual-to-visual knowledge distillation. ViLD consists of two branches: the ViLD-text branch and the ViLD-image branch.
In the former branch, fixed text embeddings obtained from VLMs' text encoder output are treated as classifiers.
Meanwhile, in the latter branch, pre-computed proposals are fed into a detector to obtain region embeddings using the RoIAlign~\cite{maskrcnn} function. The cropped image is then sent to the VLMs' image encoder to generate image embeddings. Subsequently, ViLD proposes distilling this information onto each Region-of-Interest (RoI) via $\mathcal L_1$ Loss. 
LP-OVOD~\cite{Pham2023LPOVODOO} extends the ViLD framework by making two main modifications. Firstly, it replaces the softmax cross entropy loss with the sigmoid focal loss~\cite{focal_loss}. Secondly, LP-OVOD introduces a new classification branch that is supervised by pseudo labels. 
HierKD~\cite{Ma2022HierKD} uses a single-stage detector and introduces a global-level language-to-visual knowledge distillation module. The module aims to narrow down performance gaps between one-stage and two-stage methods. This technique employs global-level knowledge distillation modules (GKD), which align global-level image representations with caption embeddings through contrastive loss.
Both ViLD and HierKD use pixel-based detectors. However, Rasheed et al.~\cite{Hanoona2022Bridging} leverages query-based detector Deformable DETR~\cite{zhu2020deformabledetr} for its detection process. 
To ensure consistency between their detection region representations and CLIP's region representations, the authors employed inter-embedding relationship matching loss (IRM). Furthermore, the authors adopt mixed datasets pre-training to enhance the ability of novel class discovery.
OADP~\cite{OADP} thinks previous works only distill object-level information from VLMs to downstream detectors and ignore the relation between different objects. 
To tackle this problem, OADP employs object-level distillation as well as global and block distillation methods. These supplementary techniques aim to compensate for the lack of relational information in object distillation by optimizing the L1 distance between the CLIP visual encoder and the detector backbone's global features or block features.
Rather than using simple novel class names for text distillation, several works also utilize more fine-grained information, including attributes, captions, and relationships of objects. 
PCL~\cite{cho2023open} adopts an image captioning model to generate more comprehensive captions that describe object instances. 
OVRNet~\cite{chen2023ovarnet} simultaneously detects objects and their visual attributes in open vocabulary scenarios. By exploring the COCO attributes dataset~\cite{patterson2016coco} via a joint co-training strategy, the authors find that the recognition of fine-grained attributes works complementary for OVD.
In summary, knowledge distillation is a common design, which effectively transfer VLM's knowledge into close set detectors. 
However, the recognition ability is still within the scope of teacher VLMs.

\noindent
\textbf{Region Text Pre-training.} Another assumption of open vocabulary learning is the availability of large-scale image text pairs, which can be easily obtained in daily life. 
Since these pairs contain large enough knowledge to cover the most novel or unseen datasets for detection and segmentation. 
Most approaches adopt web-scale caption data for pre-training, which contains millions of image text pairs.
The learning of region text alignment maps the novel classes of visual features and text features into an aligned feature space.
Once trained for the alignment, it is nature to generalize the detector for novel class classification.
OVR-CNN~\cite{zareian2021opendet_ovrcnn} first introduces the concept of open vocabulary object detection by using caption data for novel class detection.  The model first trains a ResNet\cite{resnet} and vision to language (V2L) layer using image-caption pairs via grounding, masked language modeling, and image-text matching. Since captions are not constrained in language space, the V2L layer learns to map features from visual space into semantic space without limiting to closed-label space. During the next stage of training, Faster R-CNN~\cite{ren2015faster} is used as a detection method with pre-trained ResNet as its backbone. OVR-CNN~\cite{zareian2021opendet_ovrcnn} replaces only learnable classifiers with fixed text embeddings from pre-trained language models~\cite{devlin2018bert}. 
For classification purposes, region visual features obtained from RoI-Align~\cite{maskrcnn} are sent into the V2L layer and mapped into semantic space.
Attribute-Sensitive OVR-CNN~\cite{Buettner2023EnhancingTR} proposes a different approach from OVR-CNN~\cite{zareian2021opendet_ovrcnn}. Instead of grounding vision regions to the input word embeddings of BERT~\cite{devlin2018bert}, Attribute-Sensitive OVR-CNN aligns vision regions with contextualized word embeddings that are output from BERT~\cite{devlin2018bert}. 
Additionally, Attribute-Sensitive OVR-CNN proposes using an adjective-noun negative caption sampling strategy to enhance the model's sensitivity to adjectives, verb phrases, and prepositional phrases other than object nouns in the caption.
GLIP series~\cite{li2021grounded,zhang2022glipv2} unify the object detection and phrase grounding for pre-training. In particular, it leverages massive image-text pairs by generating grounding boxes in a self-training fashion, which sets strong results for both detection and grounding.
RegionCLIP~\cite{Zhong2021RegionCLIPRL} learns visual region representation by matching image regions to region-level descriptions.
It creates pseudo labels by CLIP for region-text pairs and then uses contrastive loss to match them before fine-tuning the visual encoder using human-annotated detection datasets.
OWL-ViT~\cite{minderer2022simple} removes the final token pooling layer of pre-trained VLMs' image encoder and attaches a lightweight classification head and box regression head to each transformer 
output token before fine-tuning it on standard detection datasets using bipartite matching loss.
Meanwhile, MaMMUT~\cite{kuo2023mammut} presents a simple text decoder and visual encoder for multimodal pre-training. 
Designing a two-pass text decoder combines both contrastive and generative learning in one framework. The former is for grounding text visual entities, while the latter learns to generate.
DITO~\cite{Kim2023DITO} presents a new image-level pretraining strategy to bridge the gap between image-level pretraining
and open-vocabulary object detection. At the pertaining phase, DITO replaces the classification architecture used in CLIP with the detector architecture, which better serves the region-level recognition needs of detection by enabling the detector heads to learn from noisy image-text pairs.
In summary, adopting more text-image pair can improve the performance on rare and novel classes. However, such process needs more computation cost for extra dataset training.

\noindent
\textbf{Training with More Balanced Data.} Rare and unseen data are common in image classification datasets. 
Joint training can be used to address this issue. The core idea of these approaches is to leverage more balanced data, including image classification datasets, pseudo labels from image-text data, extra-related detection data, or even data generated by generation models.
Detic~\cite{zhou2022detecting} improves long-tail detection performance with image-level supervision. 
The classification head of Detic is trained using image-level data from ImageNet21K~\cite{russakovsky2015imagenet}. During training, the max area proposal from RPN is chosen as the RoI. 
Detic's basic insight is that most classification data are object-centric. Therefore, the maximum area proposal may completely cover an object represented by an image-level class.
The mm-ovod~\cite{Kaul2023MultiModalOVOD} method improves upon Detic~\cite{zhou2022detecting} by utilizing multi-modal text embeddings as the classifier. This approach employs a large language model to create a description of each class to generate text-based embeddings. 
In addition, mm-ovod also uses vision-based embeddings from image exemplars. By fusing the text-based and vision-based embeddings, the multi-modal text embeddings can significantly enhance Detic's performance.
To use more data, several methods generate pseudo bounding box annotations from large-scale image-caption pairs. 
PB-OVD~\cite{Gao2021pbovd} uses Grad-CAM~\cite{Selvaraju2016GradCAMVE} and generates the pseudo-bounding boxes combined with RPN's proposals. 
The activation map of Grad-CAM is obtained from the alignment between region embeddings and word embeddings that come from pre-trained VLMs.
Then, the boxes are generated from the activation map and jointly trained with existing box annotations.
Meanwhile, several works leverage the rich semantics available in recent vision and language models to localize and classify objects in unlabeled images.
VL-PLM~\cite{zhao2022VLPLM} proposes to train Faster R-CNN as a two-stage class-agnostic proposal generator using a detection dataset without category information. 
LocOV~\cite{Bravo2022LocalizedVM} uses class-agnostic proposals in RPN to train Faster R-CNN by matching the region features and word embeddings from image and caption, respectively.
From the data generation view, several works~\cite{zhao2022xpaste,li2023guiding,MosaicFusion} adopt the diffusion model to generate the on-target data for effective training. 
In particular, X-Paste~\cite{zhao2022xpaste} generates the rare class data to improve the classification ability for existing approaches. 
It uses an extra segmentation model to provide the foreground object masks and adopt simple copy and paste~\cite{ghiasi2021simple} to augment training data. 
Recently, several works~\cite{owlv2,arandjelovic2023threeway,dst_ov_det} have explored the self-training approaches to generate large pseudo labels for more balanced learning.
OWLv2~\cite{owlv2} presents a self-training pipeline. It generates huge pseudo-box annotations on WebLI~\cite{chen2022pali} dataset and pre-trains a model on such generated dataset. Finally, it fine-tunes this model on a specific OVD dataset. Since the WebLI contains a large vocabulary size, OWLv2 achieves significant gains on LVIS and COCO datasets.
In summary, these approaches are more effective in unbalanced data setting. Designing a more efficient data augmentation still have room to explore.

\noindent
\textbf{Prompting Modeling.} Prompt modeling is an effective technique for adapting foundation models to various domains, such as language modeling~\cite{brown2020language_gpt3} and image classification~\cite{zhang2021tip,zhou2022coop}. 
By incorporating learned prompts into the foundation model, the model can transfer its knowledge to downstream tasks more easily.
To generate text embeddings of category names, prompts are fed to the text encoder of pre-trained VLMs. However, negative proposals do not belong to any specific category.
To address this issue, DetPro~\cite{Du2022LearningTP} forces the negative proposal to be equally dissimilar to any object class instead of using a background class. 
PromptDet~\cite{feng2022promptdet} introduces category descriptions into the prompt and explores the position of the category in the prompt. 
It also proposes to use cached web data to enhance the novel classes during training.
Based on the DETR~\cite{detr} framework, CORA~\cite{wu2023cora} proposes region prompting and anchor pre-matching. 
The former reduces the gap between the whole image and region distributions by prompting the region features of the CLIP-based region classifier, while the latter learns generalizable object localization via a class-aware matching mechanism.
Prompt-OVD~\cite{song2023prompt} follows the pipeline of OV-DETR~\cite{zang2022open}. 
It presents RoI-based masked attention and RoI pruning techniques by utilizing CLIP visual features to improve the novel object classification.
Despite the effectiveness, without more data training, the performance of prompting is limited compared with other directions.

\noindent
\textbf{Region Text Alignment.} Using language as supervision instead of a ground truth bounding box is an attractive alternative for open vocabulary object detection. However, obtaining enough object-language annotations is difficult and costly.
Compared with region text pre-training, region text alignment aims at a better matching between region visual features and text features during the base class training without introducing extra data.
OV-DETR\cite{zang2022open} introduces a transformer-based detector for open vocabulary object detection by replacing the bipartite matching method with a conditional binary matching mechanism.
VLDet\cite{VLDet} converts the image into a set of regions and the caption into a set of words, and solves the object-language alignment problem using a set matching method. 
It proposes a simple matching strategy to align caption and vision features.
DetCLIPv2~\cite{DetCLIPv2} uses ATSS~\cite{atss} as an object detector and trains it with three datasets: a standard detection dataset, a grounding dataset, and an image-text pairs dataset for word-text alignment.
Recently, BARON~\cite{wu2023baron} proposes to align the embedding in bags of different regions rather than only individual regions. It first groups contextually interrelated regions as a bag and treats each region in the bag as a word in a sentence. Then, it sends the bag of regions into the text encoder to get bag-of-regions embeddings. 
These bag-of-regions embeddings will align with cropped region embeddings from the image encoder of VLMs. 
CoDet~\cite{ma2023codet} reformulates the region-word alignment as a co-occurring object discovery problem and aligns the co-occurring objects with the shared concept.
F-VLM~\cite{Kuo2022FVLMOO} finds that the origin CLIP features already have grouping effects.
It is a two-branch method similar to ViLD-text~\cite{gu2021open_vild}.
F-VLM uses a CLIP vision encoder as the backbone and applies the VLM feature pooler on the region features from the backbone to get VLM predictions.
The final result of F-VLM combines the detection scores and the VLM predictions. 
RO-ViT~\cite{Ro_ViT} builds upon F-VLM~\cite{Kuo2022FVLMOO}, which believes that the difference in position embeddings between image-level and region-level is responsible for the gap between vision-language pre-training and open vocabulary object detection. 
To address this issue, RO-ViT proposes a cropped positional embedding module in VLM that bridges the gap between vision-language pre-training and downstream open vocabulary object detection tasks.
In summary, region-text alignment is a core research topic, and it still has room to explore when considering recent large language models~\cite{touvron2023llama}.

\begin{table*}[!t]
   \centering
    \caption{Feature summarization of current open vocabulary segmentation approaches.}\vspace{-3mm}
   \scalebox{0.80}{{\begin{tabular}{c | c c }
      \toprule[0.15em]
        Method & Common Feature   \\
        \midrule[0.15em]
        \rowcolor{gray!15} Utilizing VLMs to Leverage Recognition Capabilities. & Design fusion or alignment methods to better fuse VLM knowledge into the existing segmenters. \\
        \rowcolor{orange!15} Learning from Caption Data. & Use extra caption data to ground the objects' visual features with caption data. \\
        \rowcolor{red!15} Generating Pseudo Labels. & Fully explore the potential of the VLM model to generate better masks to train the segmentation models \\ 
       \rowcolor{yellow!15} Training without Pixel-Level Annotations.  &  Combine the VLM and different unsupervised approaches to perform unsupervised mask generation or segmentation. \\
        \rowcolor{orange!15} Jointly Learning Several Tasks. & Pre-train on large-scale and easily available text-region pairs datasets or detection using a unified detection and segmentation model.  \\
         \rowcolor{pink!15} Adopting Denoising Diffusion Models. &  Explore the feature representation of text-to-image diffusion models or utilize the generation ability to augment masks. \\
      \bottomrule
   \end{tabular}}}
   \label{tab:ovs_summary}
   \vspace{-3mm}
\end{table*}

\subsection{Open Vocabulary Segmentation}
\label{sec:open_voc_seg}


Although segmentation problems can be defined in various ways, such as semantic segmentation or panoptic segmentation, we categorize the methods based on their technical aspects.

\noindent
\textbf{Utilizing VLMs to Leverage Recognition Capabilities.}
VLMs have shown remarkable performance in image classification by learning rich visual and linguistic representations. Therefore, it is natural to extend VLMs to semantic segmentation, which can be seen as a dense classification task.
For instance, LSeg~\cite{Language-driven-semantic-segmentation} aligns the text embeddings of category labels from a VLM language encoder with the dense embeddings of the input image.
This enables LSeg to leverage the generalization ability of VLMs and segment objects that are not predefined but depend on the input texts.
Following LSeg, several works propose methods to utilize the VLMs for open vocabulary segmentation tasks. 
Fusioner~\cite{Fusioner} utilizes self-attention operations to combine visual and language features in a transformer-based framework at an early stage.
ZegFormer~\cite{Decoupling-zero-shot-semantic-segmentation} decouples the problem into a class-agnostic segmentation task and a mask classification task. It uses the label embeddings from a VLM to classify the proposal masks and applies the CLIP-vision encoder to obtain language-aligned visual features for them.
After that, more approaches are proposed to extract the rich knowledge in VLMs.
SAN~\cite{side-adapter} attaches a lightweight side network to the pre-trained VLM to predict mask proposals and classification outputs. 
CAT-Seg~\cite{cat-seg} jointly aggregates the image and text embeddings of CLIP by fine-tuning the image encoder. 
Han et al.~\cite{global-knowledge-calibration} develop an efficient framework that does not rely on the extra computational burden of the CLIP model. 
OPSNet~\cite{embedding-modulation} proposes several modulation modules to enhance the information exchange between the segmentation model and the VLMs.
The modulation modules fuse the knowledge of the learned vision model and VLM, which improves the zero-shot performance in novel classes.
MaskCLIP~\cite{panoptic-MaskCLIP} inserts Relative Mask Attention (RMA) modules into a pre-trained CLIP model. It can utilize the CLIP features more efficiently.
TagCLIP~\cite{tagclip} proposes a trusty token module to explicitly predict pixels that contain objects (both base and novel) before classifying each pixel, which avoids the problem that models tend to misidentify pixels into novel categories.
Recently, several works~\cite{yu2023fcclip,omgseg,ov-seg-sem-clipdenoiser} directly fuse the frozen CLIP visual encoder. They fuse the learned CLIP score and prediction score to achieve better close-set and open-set recognition ability trade-offs.
Same as the open vocabulary detection, there are still several improve space when adopting more advanced VLMs.

\noindent
\textbf{Learning from Caption Data.}
Image caption data can provide weak supervision for identifying novel classes, as they expose potential novel category names.
Like such approaches in open vocabulary object detection~\cite{zareian2021opendet_ovrcnn}, image caption data is also explored in open vocabulary segmentation tasks also explore image caption data.
OpenSeg~\cite{OpenSeg} applies a region-word grounding loss to directly ground objects and nouns in the caption data.
CGG~\cite{CGG} combines caption grounding and caption generation losses to fully exploit the knowledge in caption data. It leverages the role of object nouns in visual grounding and the mutual benefits of words in caption generation.

\noindent
\textbf{Generating Pseudo Labels.} Intuitively, providing the model with more data on novel categories can improve classification performance. 
MaskCLIP+~\cite{Denseclip} modifies the CLIP-vision model by replacing the last pooling layer with a convolution layer, which produces dense feature maps. 
These feature maps are then used to generate pseudo labels for training a segmentation model.
OVSeg~\cite{Mask-adapted-clip} matches the proposed image regions with nouns in captions using CLIP to generate pseudo annotations.
It also proposes a mask prompt tuning module to help CLIP adapt to masked images without changing their weights.
XPM~\cite{XPM} follows a similar pseudo data generation procedure by aligning words in captions with regions in images to generate pseudo instance mask labels.
However, these approaches are limited by the data annotations, since both mask and region caption are hard to collection. Maybe more automatic data generation~\cite{diffusion_sem, kirillov2023segment} can be used in the future.

\noindent
\textbf{Training without Pixel-Level Annotations.} Despite the absence of annotations on novel objects, most methods still require base mask annotations during training, which are costly and labor-intensive to obtain.
To reduce the annotation burden, many recent works explore training segmentation models with merely weak supervision, such as image caption. GroupViT~\cite{Groupvit} proposes a semantic segmentation framework that leverages a grouping mechanism to automatically merge image patches with the same semantics. It trains the model with a contrastive loss on image-caption pairs and does not need pixel-level annotations.
PACL~\cite{PACL} enhances the contrastive loss with a patch alignment objective that aligns the image patches and the CLS token of the captions. 
ViL-Seg~\cite{Vil-Seg} combines both contrastive loss and clustering loss. SegCLIP~\cite{SegCLIP} further introduces a reconstruction loss and a superpixel-based KL loss. To compute the superpixel-based KL loss, it uses an unsupervised graph-based segmentation model~\cite{GraphSegmentation} to generate pseudo labels.
TCL~\cite{TCL} proposes a finer-grained contrastive loss, namely text-grounded contrastive loss, which explicitly aligns captions and regions. 
The model can segment the region that corresponds to a given text expression during inference.
OVSegmentor~\cite{OVSegmentor} introduces masked entity completion and cross-image mask constituency tasks to improve the training efficiency. 
Mask-free OVIS~\cite{mask-free-OVIS} generates pseudo mask annotations with purely image-text pair data to assist the segmentation models.
Since these methods are trained without mask supervision, the quality of segmentation results is low which make it hard to use in real application.

\noindent
\textbf{Jointly Learning Several Tasks.} Open vocabulary segmentation encompasses different tasks, including OVSS, OVIS, and OVPS. How to jointly learn multiple segmentation tasks in one model becomes a practical problem.
X-Decoder~\cite{X-decoder} proposes a framework that can handle various tasks, including open vocabulary semantic segmentation, open vocabulary instance segmentation, and open vocabulary panoptic segmentation.
It utilized a query-based segmentation architecture, pre-trains the model on a mixture of segmentation data and image-text pairs, and then fine-tuned or applied in zero-shot settings for downstream tasks. 
FreeSeg~\cite{freeseg} also proposes a generic framework to tackle the three tasks in a unified manner. 
It designs an adaptive task prompt module and performs test time tuning on the learnable prompts to capture the task-specific features.
POMP~\cite{ren2023prompt} first trains class prompts at a large vocabulary dataset (Imagenet-21K) and then transfers the prompts into multiple open vocabulary tasks.
Moreover, if the model can learn from multiple tasks, it raises the question of whether it can benefit from multi-sourced data, e.g., detection and segmentation data.
To handle this question, OpenSeeD~\cite{OpenSeeD} and OpenSD~\cite{ov-seg-unify-opensd} jointly learn from segmentation and detection datasets. To bridge the task gap, they propose decoupled decoding frameworks, which decode foreground and background masks separately and generate masks for the bounding box proposals.
One shortcoming of these approaches is extra computation costs that brought other tasks.

\noindent
\textbf{Adopting Denoising Diffusion Models.} 
Recently, diffusion-based generative models~\cite{ldm} have achieved remarkable success in text-based image generation suggests. There are mainly two ways to let diffusion models enhance open vocabulary tasks. First, The reality and diversity of the images generated by diffusion models suggest that the intermediate representations in the diffusion models may be highly aligned with natural language vocabularies. 
Inspired by this, ODISE~\cite{ODISE} proposes a framework that leverages the middle representation of the diffusion model. The middle representations contain rich semantic information for both base and novel classes. Thus, ODISE can perform open vocabulary segmentation by training a decoder head on the representations. \cite{ov-ins-diffcamouflage} follows ODISE to use the intermediate representations of diffusion models.
Another way to incorporate diffusion models into the open vocabulary setting is to take advantage of the image and mask generation ability~\cite{wu2023diffumask,diffusion_sem,li2023grounded}. OVDiff~\cite{diffusion_sem} proposes a prototype-based method to tackle open vocabulary semantic segmentation. It uses the diffusion model to generate images for various categories and treat them as prototypes. The method does not need training. During testing, the input images are compared with these generated prototypes, and the best-matching one is the predicted class. Meanwhile, several works~\cite{wu2023diffumask,li2023grounded} use the diffusion model to generate images and masks for the rare classes to augment data. Thus, the segmenter can be trained in a more balanced manner or even totally using generated data.
Despite the inference time of these methods is quite limited, it still has room to explore joint generation and segmentation in one framework. 

\begin{table*}[ht]
    \centering
    \tiny
    \caption{Representative works summarization and comparison in Sec.~\ref{sec:open_ov_videos}, Sec.~\ref{sec:open_voc_3d} and Sec.~\ref{sec:closely_related_tasks}. \textbf{OV-VC} refers to open vocabulary video classification. \textbf{OV-3D} refers to open vocabulary 3D recognition.}\vspace{-3mm}
    \setlength{\tabcolsep}{1.2pt}
    \scalebox{1.0}{
        \begin{tabular}{p{0.10\textwidth}p{0.10\textwidth}p{0.15\textwidth}p{0.10\textwidth}p{0.20\textwidth}p{0.30\textwidth}}
            \toprule
            \belowrulesepcolor{gray!30!}
        \rowcolor{gray!30!} Method & Task & Vision Training Annotations & Text Model & Vision Model & \ \ ~~~~~~~~~~~~~~~~~~~~~~~~~~~~~~~~~~~~Highlight \\ \aboverulesepcolor{gray!30!} \midrule
        \belowrulesepcolor{orange!15!}
        \rowcolor{orange!15!}\multicolumn{6}{c}{\textbf{Open Vocabulary Video Understanding (Sec.~\ref{sec:open_ov_videos})}} \\ \aboverulesepcolor{orange!15!} \midrule
        ActionCLIP~\cite{wang2021actionclip} & OV-VC & None & CLIP-text & CLIP-vision + Temporal Pooling & The first work proposes to use CLIP for video understanding. \\ 
        I-VL~\cite{ju2022prompting} & OV-VC  & Video Class Label (base) & CLIP-text & CLIP-vision + transformer (temporal fusion) & I-VL provides a simple baseline for adapting CLIP to video understanding. \\
        ViFi-CLIP~\cite{rasheed2022fine} & OV-VC & Video Class Label (base) & CLIP-text & CLIP-vision (learnable) + Temporal Pooling & ViFi-CLIP indicates that a simple fine-tuning baseline is strong enough. \\ 
        OVTrack~\cite{li2023ov} & OV-Tracking & Image Bounding Boxes & CLIP-text & CLIP-vision + Faster R-CNN & OVTrack is the first open vocabulary multi-object tracking method. \\
        MindVLT~\cite{wang2023towards} & OV-VIS & Video Segmentation Masks (base) & CLIP-text & CLIP-vision + Mask2Former + SORT    & MindVLT is the first open vocabulary video instance segmentation method. \\
        OpenVIS~\cite{guo2023openvis} & OV-VIS & Video Segmentation Masks (base) & CLIP-text & CLIP-vision + Mask2Former & OpenVIS conduct open vocabulary VIS in a two-stage manner.\\
        \midrule\belowrulesepcolor{orange!15!}
        \rowcolor{orange!15!}\multicolumn{6}{c}{\textbf{Open Vocabulary 3D Scene Understanding (Sec.~\ref{sec:open_voc_3d})}} \\ \aboverulesepcolor{orange!15!} \midrule
        PointCLIP~\cite{zhang2022pointclip} & OV-3D & None & CLIP-text & CLIP-vision (projected depth map)  & PointCLIP uses projected depth for OV 3D recognition. \\
        PointCLIPV2~\cite{zhu2022pointclip} & OV-3D & None & CLIP-text + GPT-3 & CLIP-vision (projected depth map) & PointCLIPV2 generates prompts with LLMs to enhance performance. \\
        ULIP~\cite{xue2023ulip} & OV-3D & Point Cloud Label & CLIP-text & Point Cloud Encoder + CLIP-vision  & ULIP aligns the 3D encoder and 2D CLIP encoder feature by distillation.\\
        CLIP2~\cite{zeng2023clip2} & OV-3D & None & CLIP-text & Point Cloud Encoder + CLIP-vision & CLIP2 leverages real-world point cloud data and CLIP to train the 3D encoder for 3D open vocabulary segmentation.\\
        OpenShape~\cite{liu2023openshape} & OV-3D & Point Cloud Label & CLIP-text + GPT-4 & Point Cloud Encoder + CLIP-vision & OpenShape builds a large-scale dataset that provides image, point cloud, and text triplets.\\
        OV-3DETIC~\cite{lu2022open}& OV-3D-OD & Pseduo Labels from 2D Detector & CLIP-text & 3DDETR~\cite{misra2021end}&  OV-3DETIC explois information from two modalities to achieve 3D open vocabulary object detection. \\
        PLA~\cite{ding2023language}& OV-3D-SS/IS & 3D segmentation masks (base) & CLIP-text & sparse-conv UNet~\cite{graham20183d} & PLA first tackles the 3D open vocabulary scene understanding problem.\\
        OpenScene~\cite{Peng2023OpenScene} & OV-3D-SS & None & CLIP-text & 3D Encoder + LSeg~\cite{Language-driven-semantic-segmentation} & OpenScene train a 3D Encoder yielding dense features co-embedded with text and image pixels for open vocabulary semantic segmentation.\\
        \bottomrule
        \end{tabular}}
\label{tab:video_3d_summary}
\vspacefigtext
\end{table*}

\subsection{Open Vocabulary Video Understanding}
\label{sec:open_ov_videos}

We also review several video open vocabulary tasks. Most works focus on designing VLM models' temporal fusion or association in various settings, including action recognition and tracking. 

\noindent
\textbf{Video Classification.}
Traditional video classification methods usually require large datasets specific to video (e.g., Kinetics~\cite{carreira2017quo}). However, annotating video datasets requires very high costs.
Using semantic information of label texts from web data may alleviate this deficiency.
Building on the success of CLIP in image open vocabulary recognition, ActionCLIP~\cite{wang2021actionclip} uses knowledge from image-based vision-language pre-training. It adds a temporal fusion layer for the zero-shot capability in video action recognition. 
A concurrent work I-VL~\cite{ju2022prompting} follows a similar paradigm by training the transformer layer on top of a frozen CLIP image encoder and also has a strong zero-shot capability. 
Another concurrent work, EVL~\cite{lin2022frozen}, also employs the frozen CLIP for efficient video action recognition. 
X-CLIP~\cite{ni2022expanding} uses a video encoder that leverages the temporal information in the encoder for video recognition. It aligns the feature from the video encoder with the text encoder that is pre-trained on the image-based vision-language pairs (e.g., CLIP). 
To take advantage of other modalities, MOV~\cite{qian2022multimodal} further fuses the audio information and the pre-trained CLIP model to build a multi-modal open vocabulary video classification model.
%
Recently, Open-VCLIP~\cite{weng2023transforming} formulates the CLIP-to-video knowledge transfer as a continual learning problem and proposes \textit{Interpolated Weight Optimization} to address the issue. 
AIM~\cite{yang2023aim} tackles the open vocabulary video classification problem by adding adapt layers on top of the CLIP image encoder. 
ViFi-CLIP~\cite{rasheed2022fine} reveals that a simple fine-tuning baseline (image-level feature extraction with CLIP visual encoder following temporal pooling) instead of advanced fusion layers can have a strong performance and further investigates the influence of prompt tuning. 
ASU~\cite{chen2023video} explores the use of fine-grained language features extracted by semantic units~(e.g., head, arms, balloon, knee bend posture of exercise in the video) to guide the training of video classification. 
Recent advancements in open vocabulary video classification have turned our attention to the language part of VLM modeling. 
VicTR~\cite{kahatapitiya2023victr} introduces \textit{video-conditioned text representations} to optimize the visual and text information jointly.
MAXI~\cite{lin2023match} leverages LLMs to build a text bag for video without annotation by text expansion. The verbs, which are lacking in the realm of image, have the opportunity to participate in video-language modeling.

\noindent
\textbf{Object Tracking and Video Instance Segmentation.} 
The object tracking and video instance segmentation can also enjoy the rich knowledge of VLMs to build an open vocabulary tracker based on a close-set tracker.
Going beyond the large vocabulary object tracking~\cite{li2022tracking}, to tackle the real-world multiple object tracking (MOT), OVTrack~\cite{li2023ov} first introduces large VLMs to the object tracking and tackles their proposed open vocabulary MOT (OV-MOT) task. 
Specifically, OVTrack extracts RoIs via an RPN and uses CLIP for knowledge distillation. 
A separate tracking head is used for tracking and supervised by pseudo-LVIS videos. 
As for video instance segmentation (VIS), to make the VIS model capable of generalizing to novel classes in the real world,
MindVLT~\cite{wang2023towards} adopts a frozen CLIP backbone and proposes an end-to-end method with an open vocabulary classifier for segmenting and tracking unseen categories.
It collects a large-vocabulary VIS dataset and tests their MindVLT on it.
A concurrent work, OpenVIS~\cite{guo2023openvis}, also tackles the open vocabulary video instance segmentation task but in a different manner. It generates the class-agnostic mask of instances and leverages the masks to crop raw images to feed into the CLIP visual encoder for calculating class scores.
Beyond the open vocabulary video instance segmentation, DVIS++~\cite{zhang2023dvispp} proposes the first open vocabulary universal video segmentation scheme supporting video semantic segmentation, video instance segmentation, and video panoptic segmentation. DVIS++~\cite{zhang2023dvispp} leverages the knowledge from the CLIP backbone and adopts mask pooling on the CLIP backbone for enabling open vocabulary capability.

\vspace{-2mm}
\subsection{Open Vocabulary 3D Understanding}
\label{sec:open_voc_3d}

In this section, we review the open vocabulary approaches used in 3D scene understanding tasks. We mainly survey the works for point cloud classification and segmentation, which explores the knowledge of 2D VLMs into 3D.

\noindent
\textbf{3D Recognition.} The VLM has shown great success in the zero-shot or few-shot learning of 2D images by leveraging a huge amount of image data. However, such internet-scale data is not available for the 3D point cloud. To this extent, extending the open vocabulary mechanism into 3D perception is not trivial since it is hard to collect enough data for point-language contrastive training like what CLIP does in 2D perception.
To mitigate the gap, PointCLIP~\cite{zhang2022pointclip} takes the first step to 3D open vocabulary perception. The principle insight behind PointCLIP is that the 3D point cloud can be converted to CLIP-recognizable images. By projecting the 3D point cloud to a 2D plane and extracting visual features from the projected depth map via the CLIP visual encoder, the point cloud feature can be naturally aligned with the language feature extracted by the language encoder. Then, the whole framework is with 3D point cloud zero-shot capability, just like 2D images.
However, simply projecting the point cloud to depth maps may yield inferior performance. To further improve the performance, rather than directly using the CLIP visual encoder for visual feature extraction of depth map, CLIP2Point~\cite{huang2022clip2point} aligns the RGB image feature from the CLIP visual encoder and depth feature from a depth encoder through contrastive learning. Specifically, they collect image-depth pairs to align the from-scratch depth map encoder with the CLIP image encoder at training time and only use the depth map encoder at inference time. Then, the depth feature can be aligned with the language embedding, facilitating the 3D point cloud zero-shot capability with CLIP.
Also aiming at improving the performance of PointCLIP~\cite{zhang2022pointclip}, PointCLIPV2~\cite{zhu2022pointclip} focuses on different perspectives and proposes a realistic shape projection scheme along with an LLM-based 3D prompting generation scheme. 
Although without extra annotations, PointCLIPV2 achieves a significant improvement over PointCLIP~\cite{zhang2022pointclip}.

Though projecting the point cloud to depth maps can make it easy to leverage the existing 2D visual encoders pre-trained in the CLIP, it may not exhaustively use the 3D information in the point cloud.
Accordingly, beyond the depth map-based methods, ULIP~\cite{xue2023ulip} first collects multi-modalities triplets (point cloud, image, and text) for training the 3D backbone. Since CLIP already aligns the language and image encoders, it only needs to align the 3D backbone to the image-language feature space. 
With only small-scale data, ULIP~\cite{xue2023ulip} aligns the 3D feature extracted by the 3D backbone to the CLIP-aligned visual and text feature to enable the zero-shot capability and enhance the standard 3D recognition capability. 
The advantage of ULIP~\cite{xue2023ulip} also includes unifying the three modalities, which may bring more downstream applications, such as image-to-3D retrieval.
Though ULIP~\cite{xue2023ulip} aligns the 3D and 2D and thus enables the native 3D open vocabulary capability, it trains on a small-scale dataset, which limits its performance.
CLIP2~\cite{zeng2023clip2} resorts to finding training samples from the real world and proposes \textit{Triplet Proxies Collection} scheme for finding instance-level 3D point cloud, 2D image crop, and text triplets for 3D recognition.
OpenShape~\cite{liu2023openshape} scales up both the dataset and backbone. For the dataset, they build a text-3D shape dataset containing 876k training shapes with over 1k categories. They also explore adopting larger 3D backbones and their performance. As a result, OpenShape~\cite{liu2023openshape} drastically enhances the performance on 3D zero-shot capability.
LidarCLIP~\cite{hess2022lidarclip} also aligns the 3D and 2D features with a similar approach, but LidarCLIP~\cite{hess2022lidarclip} focuses on the driving scenes. 
Besides the 3D zero-shot capability, LidarCLIP~\cite{hess2022lidarclip} also shows point cloud captioning and lidar-to-image generation capability with off-the-shelf foundation models.

\noindent
\textbf{3D Object Detection.} 
In 3D object detection, the training data is often hard to get, and existing 3D detectors are often trained on very limited classes, which means they are hard to generalize to novel classes.
Inspired by the success of 2D open vocabulary detection~\cite{zhou2022detecting}, aiming at taking full use of both image and point cloud modalities for 3D detection, OV-3DETIC~\cite{lu2022open} and OV-3DET~\cite{lu2023open} proposes to decouple the localization and recognition in point cloud object detection into localization and recognition. For localization, 2D pre-trained detectors can be adapted to train the 3D detector by back-projecting the 2D bounding box and further optimizing it by the point cloud. 
For recognition, they propose aligning the regional features of 3D and 2D encoders with the corresponding text feature. Therefore, during the inference, 3DETIC~\cite{lu2022open} and OV-3DET~\cite{lu2023open} can generalize to unseen classes with the knowledge in the CLIP. While OV-3DET~\cite{lu2023open} achieves significant performance on the Open-Vocabulary 3D detection, it relies on the localization capability of pre-trained 2D detectors, which may limit the 3D object discovery capability. To solve this problem, CODA~\cite{cao2023coda} proposes to discover objects with both 2D semantic prior and 3D geometry prior, and further proposes a cross-modal alignment module to align 3D and 2D features. Also trying to get rid of the 2D image detector, concurrent work Object2Scene~\cite{zhu2023object2scene} proposes to augment the existing 3D object detection datasets with large-scale 3D object datasets. 
It also introduces a new framework, L3Det~\cite{zhu2023object2scene}, for 3D object-text alignment. As a result, both CODA~\cite{cao2023coda} and L3Det~\cite{zhu2023object2scene} outperforms OV-3DET~\cite{lu2023open}. As driving scenarios may yield new challenges, the above-mentioned methods may have an inferior performance. 
OpenSight~\cite{zhang2023opensight} focuses on the outdoor scenes and proposes to leverage the knowledge from 2D grounding DINO~\cite{liu2023groundingdino} to train the 3D detectors. In particular, it uses the size prior (e.g., cars are 1.96m wide) generated by the LLM to re-calibrate the output of the model to make the model more suitable for the driving scene.

\noindent
\textbf{3D Scene Understanding.}
In the understanding of the 3D scene, the same obstacle, lacking training data, also limits the capability of generalization. 
Sharing a similar motivation that there is a lack of 3D-text pairs for point-language contrastive training, PLA~\cite{ding2023language} proposes to extract features from multi-view images sampled from a scene to generate descriptions with pre-trained language models. 
The descriptions can then be used to extract language features to train 3D backbone-language alignment. With the language-aligned 3D backbone, it is natural to conduct open vocabulary segmentation like with the 2D image.
OpenScene~\cite{Peng2023OpenScene} proposes to link the 3D features for each point with CLIP features for each pixel. By back-projecting the 2D pixels to 3D space, each point in the point cloud can be ensembled with features of several pixels in different views. The 3D-text co-embedding makes it possible for open vocabulary semantic segmentation.
Inspired by MaskCLIP~\cite{Denseclip}, a concurrent work CLIP-FO3D~\cite{zhang2023clip} also focuses on 3D semantic segmentation and has a similar approach. It leverages the feature map extracted by the CLIP visual encoder and directly transfers CLIP’s knowledge without any extra annotation.
A recent work, RegionPLC~\cite{yang2023regionplc} on 3D open vocabulary semantic segmentation, considers the benefits of both PLA~\cite{ding2023language} and OpenScene~\cite{Peng2023OpenScene}. 
It proposes to use the dense caption of any random region to extract more information in CLIP for 3D backbone distillation. Fine-grained dense supervision in RegionPLC~\cite{yang2023regionplc} beyond view-level or instance-level further improves performance compared to PLA~\cite{ding2023language}.
Different from training 3D backbones like in~\cite{ding2023language, Peng2023OpenScene, zhang2023clip, yang2023regionplc}, PartSLIP~\cite{liu2023partslip} and SATR~\cite{abdelreheem2023satr} directly projecting the 3D point cloud and mesh to 2D plane and uses GLIP~\cite{li2021grounded} to localize and segment.
Meanwhile, OpenMask3D~\cite{takmaz2023openmask3d} mainly focuses on the 3D instance segmentation. It first generates class-agnostic instance mask proposals with the 3D point cloud. Then, OpenMask3D~\cite{takmaz2023openmask3d} selects views and projects 3D masks into 2D images. The 2D masks are further refined by SAM~\cite{kirillov2023segment}. Finally, the 2D masks can be fed into the CLIP visual encoder to generate label prediction with the help of the CLIP language encoder.
While OpenMask3D~\cite{takmaz2023openmask3d} has a strong performance, it requires point cloud-2D image pairs for training. On the contrary, OpenIns3D~\cite{huang2023openins3d} does not require 2D images but only needs point clouds with colors. OpenIns3D~\cite{huang2023openins3d} proposes a "Mask-Snap-Lookup" pipeline that first learns class-agnostic mask proposals, then generates synthetic scene-level images, and finally assigns categories for each proposal.
Another work trying to improve OpenMask3D~\cite{takmaz2023openmask3d} is Open3DIS~\cite{nguyen2023open3dis}. Open3DIS~\cite{nguyen2023open3dis} improves the 3D mask proposal quality by employing a 3D instance network and a 2D-guide-3D Instance Proposal Module. After the mask proposal, it aggregates CLIP features in a multi-scale, multi-view manner. Open3DIS~\cite{nguyen2023open3dis} achieves a stronger performance over OpenMask3D~\cite{takmaz2023openmask3d} on several datasets.
In the driving scene, to mitigate the heavy dependence on the point cloud data annotation and fast generalization to the new scenes, CLIP2Scene~\cite{chen2023clip2scene} proposes to train a 3D network via semantic-driven cross-modal contrastive learning for 3D segmentation. CLIP2Scene~\cite{chen2023clip2scene} also considers spatial-temporal semantic consistency regularization to facilitate training.

\subsection{Closely Related Tasks}
\label{sec:closely_related_tasks}


\noindent$\bullet$
\textbf{Class Agnostic Detection and Segmentation.} The goal of class agnostic detection and 
segmentation is to learn a general region proposal system that can be used in different scenes. 
For detection, OLN~\cite{kim2021oln} replaces the binary classification head in RPN by predicting the IoU score of foreground objects, which proves the generalization ability on cross-dataset testing. For example, the model trained on COCO shows the localization ability on the LVIS dataset.
Open world instance segmentation~\cite{wang2021unidentified} aims to correctly detect and segment instances whether their categories are in training taxonomy.
GGN~\cite{wang2022ggn} leverages the bottom-up grouping idea, combining a local pixel affinity measure with instance-level mask supervision, producing a training regimen designed to make the model more generalizable.
Meanwhile, from the data augmentation view, MViT~\cite{Maaz2022Multimodal} combines the Deformable DETR and CLIP text encoder for the open world class-agnostic detection, where the authors build a large dataset by mixing existing detection datasets. 
For the segmentation domain, Entity segmentation~\cite{qi2022openentity,qi2022fineopenentity} aims to segment all visual entities without predicting their semantic labels. The goal of such segmentation is to obtain high-quality and generalized segmentation results. 
Fined-grained entity~\cite{qi2022fineopenentity} presents a two-view image crop ensemble method named CropFormer to enhance the fine-grained details. 
Recently, SAM~\cite{kirillov2023segment} proposes more generalized prompting methods, including masks, points, boxes, and texts. 
In particular, the authors build a larger dataset with 1 billion masks following the spirits of CLIP.
The SAM achieves zero-shot testing in various segmentation datasets, which are highly generalizable.

\noindent$\bullet$
\textbf{Open World Object Detection.} Open world recognition~\cite{openworld} requires the model to identify novel categories and label them as ``unknown.'' Then, the novel categories are progressively annotated.
The model learns incrementally with new data and recognizes the newly-annotated categories.
Inspired by this setting, open world object detection (OWOD)~\cite{openworld-detection} expands the recognition task to object detection. 
The authors propose a novel method that utilizes contrastive clustering and an energy-based unknown identification module. 
Following them, OW-DETR~\cite{OW-DETR} introduces an end-to-end transformer-based framework. 
It consists of three dedicated components, namely, attention-driven pseudo-labeling, novelty classification, and objectness scoring, to address the OWOD challenge explicitly. 
Meanwhile, open world DETR~\cite{Open-World-DETR} proposes a two-stage training approach based on Deformable DETR. 
It focuses on alleviating catastrophic forgetting when the annotations of the unknown classes become available incrementally using knowledge distillation and exemplar replay technologies. 
PROB~\cite{PROB} proposes a probabilistic objectness head into the OW-DETR to better mine unknown background classes.
Recently, researchers~\cite{UC-OWOD} propose unknown-classified open world object detection (UC-OWOD), which aims to detect and classify unknown instances into different unknown classes. This task is close to the open vocabulary object detection but requires incremental learning.

\noindent$\bullet$
\textbf{Open-Set Panoptic Segmentation.} Similar to other open-set tasks, open-set panoptic segmentation (OSPS)~\cite{EOPSN} requires the model to identify novel categories as `unknown' in a panoptic segmentation task. 
The 'un classes are chosen from the thing classes (foreground objects).
The authors apply exemplar theory for novel class discovery. 
After that, Dual~\cite{dual} proposes a divide-and-conquer scheme to develop a dual decision process for OSPS. 
The results indicate that by properly combining a known class discriminator with an additional class-agnostic object prediction head, the OSPS performance can be significantly improved.

\section{CHALLENGES AND OUTLOOK}
\label{sec:future_direction}


\subsection{Challenges} 
\label{subsec:challenges}

\noindent
\textbf{Base Classes Over-fitting Issues.} Most approaches detect and segment novel objects by learning proposals from base class annotations. Thus, there are natural gaps in shapes and semantic information between novel and base objects. VLM models can bridge such gaps via pre-trained visual-text knowledge. However, most detectors still easily overfit the base classes when the novel classes have similar shapes and semantics, since these classes are trained with higher confidence scores. More fine-grained feature discriminative modeling~\cite{peize2023vlpart,qi2023aims,pan2023towards}, including parts or attributes, is required to handle these problems.

\noindent
\textbf{Training Costs.} Most state-of-the-art methods need \textit{huge} data for pre-training to achieve good performances. However, the costs are expensive and even unavailable for many research groups to follow. Thus, with the aid of VLM, designing more efficient data learning pipelines or learning methods is more practical and affordable. One simple solution can be adopting a frozen backbone. However, this may limit the representation capacity.

\noindent
\textbf{Across Dataset Generation and Evaluation.} As shown in the benchmark section, current state-of-the-art methods design specific models for each benchmark. Designing one shared model~\cite{liu2023groundingdino,zhao2024open_groundingdino,omgseg, li2023semantic, yuan2024ovsam} across different datasets on open vocabulary detection and segmentation follows the origin spirits of open vocabulary learning.
However, there are still performance gaps between unified models and dataset-specific models in several OVD benchmarks, such as LVIS datasets.

\noindent
\textbf{Better Benchmarks and Metrics.} Since several classes contain overlapping concepts (for example, building and tower, person and woman), designing more new metrics~\cite{zhou2023rethinking} is also needed to better measure the open vocabulary methods. Moreover, current datasets are still small. To realize real open vocabulary settings, more datasets like SAM-1B~\cite{kirillov2023segment} are needed.

\subsection{Future Work}
\label{subsec:future_work}

\noindent
\textbf{Explore Temporal Information.} In practical applications, video data is readily available and used more frequently. Accurately segmenting and tracking objects that are not predetermined requires a great deal of attention, which is necessary for a wide range of real-world scenarios, such as short video clips and autonomous vehicles. However, there are only a few works~\cite{li2023ov} exploring open vocabulary learning on detection and tracking in video. Moreover, the input scenes~\cite{wang2023towards} are simple. For example, several clips only contain a few instances, which makes the current tracking solution more trivial. Thus, a more dynamic, challenging video dataset is needed to fully explore the potential of vision language models for open vocabulary learning.

\noindent
\textbf{3D Open Vocabulary Scene Understanding.} Compared with image and video, point cloud data are more expensive to annotate, in particular for dense prediction tasks. Thus, research on 3D open vocabulary scene understanding is more urgent. Current solutions for 3D open vocabulary scene understanding focus on designing projection functions for better usage of 2D VLMs. More new solutions for aligning 2D models knowledge into 3D models will be a future direction.

\noindent
\textbf{Explore Foundation Models With Specific Adapter For Custom Tasks.} Vision foundation models~\cite{kirillov2023segment,CLIP} can achieve good zero-shot performance on several standard classification and segmentation datasets. However, for several custom tasks, such as medical image analysis, and aerial images, there are still many corner cases. Thus, designing a task-specific adapter is needed for these custom tasks. Such adapters can fully utilize the knowledge of pre-trained foundation models. One possible solution is to explore the in-context learning~\cite{fang2023explore,Painter} to fully explore or connect the knowledge of VLMs and LLMs.

\noindent
\textbf{Combining with Incremental Learning.} In real scenarios, the data annotations are usually open-world and non-stationary, where novel classes may occur continuously and incrementally. However, directly turning into incremental learning may lead to catastrophic forgetting problems. On the other hand, current open world object detection~\cite{OW-DETR} only focuses on novel class localization, rather than classification. How to handle both catastrophic forgetting problems and novel class detection in one framework is worth exploring in the future.

\noindent
\textbf{Combining with Large Language Models.} Compared with VLMs, most LLMs contain more text concepts, which naturally have a broader scope than various dataset taxonomies, even larger than the recent V3Det~\cite{wang2023v3det} dataset. Thus, how to better align the LLMs knowledge~\cite{liu2023llava} with visual detectors or segmenters to achieve stronger zero-shot results still needs exploration.

\section{Conclusion}
\label{sec:conclusion}
This survey offers a detailed examination of the latest developments in open vocabulary learning in computer vision, which appears to be a first of its kind. 
We provide an overview of the necessary background knowledge, which includes fundamental concepts and introductory knowledge of detection, segmentation, and vision language pre-training. 
Following that, we summarize more than 50 different models used for various scene understanding tasks. For each task, we categorize the methods based on their technical viewpoint. 
Additionally, we provide information regarding several closely related domains. 
In the experiment section, we provide a detailed description of the settings and compare results fairly. 
Finally, we summarize several challenges and also point out several future research directions for open vocabulary learning.

\noindent
\textbf{Acknowledgement.} This work is supported by the National Key Research and Development Program of China (No. 2023YFC3807600) and the interdisciplinary doctoral grants (iDoc 2021-360) from the Personalized Health and Related Technologies (PHRT) of the ETH domain.
\appendix

\begin{table*}[t!]
  \centering
  \caption{\textbf{Open vocabulary video classification results.} The mean and standard deviation of results are reported.}
  \begin{adjustbox}{width=1.\textwidth}
    \begin{tabular}{lcccccccc} 
      \toprule[0.15em]
      Method  & \makecell{Vision Backbone}  & \makecell{Frames}   & \makecell{Ground Truth}  & UCF101 & HMDB51 & K600 (Top-1) & K600 (Top-5)  \\
      \toprule[0.15em]
        ActionCLIP~\cite{wang2021actionclip}\textit{(Arxiv '2109)} & ViT-B/16 & 32 & \cmark & 58.3 $\pm$ 3.4 & 40.8 $\pm$ 5.4 & 66.7 $\pm$ 1.1 & 91.6 $\pm$ 0.3 \\
        I-VL~\cite{ju2022prompting}\textit{(ECCV '22)} & ViT-B/16 & 32 & \cmark & 69.3 $\pm$ 4.2 & 44.3 $\pm$ 2.2 & 55.8 $\pm$ 0.7 & 81.4 $\pm$ 0.3 \\
        X-CLIP~\cite{ni2022expanding}\textit{(ECCV '22)} & ViT-B/16 & 32 & \cmark &   72.0 $\pm$ 2.3 & 44.6 $\pm$ 5.2 & 65.2 $\pm$ 0.4 & 86.1 $\pm$ 0.8 \\
        Text4Vis~\cite{wu2023revisiting}\textit{(AAAI '23)} & ViT-L/14 & 16 & \cmark & - & - & 68.9 $\pm$ 1.0 & -\\
        ViFi-CLIP~\cite{rasheed2022fine}\textit{(CVPR '23)} & ViT-B/16 & 32 & \cmark & 76.8 $\pm$ 0.7 & 51.3 $\pm$ 0.6 & 71.2 $\pm$ 1.0 & 92.2 $\pm$ 0.3 \\
        ASU~\cite{chen2023video}\textit{(Arxiv '2303)} & ViT-B/16 & 32 & \cmark & 75.0 $\pm$ 3.7 & 48.1 $\pm$ 2.8 & 67.6 $\pm$ 0.2 & 87.2 $\pm$ 0.3\\
        MAXI~\cite{lin2023match}\textit{(Arxiv '2303)} & ViT-B/16 & 32 & \xmark & 78.2 $\pm$ 0.8 & 52.3 $\pm$ 0.7 & 71.5 $\pm$ 0.8 & 92.5 $\pm$ 0.4   \\
        VicTR~\cite{kahatapitiya2023victr}\textit{(Arxiv '2304)} & ViT-B/16 & 32 & \cmark & 72.4 $\pm$ 0.3 & 51.0 $\pm$ 1.3 & -\\
        Open-VCLIP~\cite{weng2023transforming}\textit{(ICML '23)}  & ViT-B/16 & 16 & \cmark & 83.4 $\pm$ 1.2 & 53.9 $\pm$ 1.2 & 73.0 $\pm$ 0.8 & 93.2 $\pm$ 0.1 \\
        Open-VCLIP~\cite{weng2023transforming}\textit{(ICML '23)}  & ViT-L/14 & 16 & \cmark & 87.6 $\pm$ 1.2 &  59.0 $\pm$ 0.6 & 81.1 $\pm$ 0.8 & 96.3 $\pm$ 0.3  \\
      \toprule[0.15em]
    \end{tabular}
  \end{adjustbox}

  \label{table:OV-VC}
\end{table*}

\begin{table*}[t]
\centering
    \caption{The performance comparison on LV-VIS~\cite{wang2023towards} validation and test set. The $AP_{all}$, $AP_{base}$, and $AP_{novel}$ mean the average precision of overall categories, base categories, and novel categories.}
    \begin{tabular}{c c c c c c c c}
        \toprule
        \multirow{2}{*}{Method}&\multirow{2}{*}{Backbone}&\multicolumn{3}{c}{Val}&\multicolumn{3}{c}{Test}\\
        \cmidrule(lr){3-5}
        \cmidrule(lr){6-8}
        && $mAP_{all}$ & $mAP_{base}$ & $mAP_{novel}$ & $mAP_{all}$ & $mAP_{base}$ & $mAP_{novel}$\\
        \midrule
        DetPro~\cite{Du2022LearningTP}-SORT~\cite{wojke2017simple}&R50&9.9&14.9&4.1&5.7&8.8&2.8\\
        Detic~\cite{zhou2022detecting}-SORT~\cite{wojke2017simple}&R50&10.1&15.1&4.2&5.7&8.5&3.1\\
        DetPro~\cite{Du2022LearningTP}-OWTB~\cite{liu2022opening}&R50&11.8&17.1&5.7&7.1&9.9&4.0\\
        Detic~\cite{zhou2022detecting}-OWTB~\cite{liu2022opening}&R50&11.5&16.8&5.3&7.1&10.1&3.8\\
        MindVLT~\cite{wang2023towards}&R50&\textbf{14.1}&\textbf{18.1}&\textbf{9.4}&\textbf{8.6}&\textbf{10.7}&\textbf{6.4}\\
        \midrule
        Detic~\cite{zhou2022detecting}-SORT~\cite{wojke2017simple}&SwinB&14.9&20.7&8.2&11.0&16.2&5.4\\
        Detic~\cite{zhou2022detecting}-OWTB~\cite{liu2022opening}&SwinB&15.8&21.5&9.1&11.7&16.4&6.6\\
        MindVLT~\cite{wang2023towards}&SwinB&\textbf{21.4}&\textbf{26.3}&\textbf{15.8}&\textbf{14.9}&\textbf{18.8}&\textbf{10.6}\\
        \bottomrule
    \end{tabular}
\label{tab:OV-VIS}
\end{table*}

\begin{table*}[t]
\scriptsize
  \centering
  \caption{The performance comparision of zero-shot classification on Objaverse-LVIS~\cite{deitke2023objaverse}, ModelNet40~\cite{wu20153d}, and ScanObjectNN~\cite{uy2019revisiting}.}
    \begin{tabular}{c|c|ccc|ccc|ccc}
    \toprule
    \multirow{2}[4]{*}{Method} & \multirow{2}[4]{*}{Training}  & \multicolumn{3}{c|}{Objaverse-LVIS~\cite{deitke2023objaverse}} & \multicolumn{3}{c|}{ModelNet40~\cite{wu20153d}} & \multicolumn{3}{c}{ScanObjectNN~\cite{uy2019revisiting}} \\
    \cmidrule{3-11}
    && Top1  & Top3  & Top5  & Top1  & Top3  & Top5  & Top1  & Top3  & Top5 \\
    \midrule
    PointCLIP~\cite{zhang2022pointclip} & \multirow{2}{*}{None} & 1.9   & 4.1      & 5.8   & 19.3  & 28.6      & 34.8  & 10.5  & 20.8      & 30.6 \\
    PointCLIP v2 \cite{zhu2022pointclip} && 4.7   & 9.5      & 12.9  & 63.6  & 77.9      & 85.0  & 42.2  & 63.3      & 74.5 \\
    \midrule
    CG3D~\cite{hegde2023clip} &   \multirow{5}{*}{ShapeNet~\cite{chang2015shapenet}}    & 5.0      & 9.5       & 11.6      & 48.7  & 60.7      & 66.5      & 42.5    & 57.3      & 60.8  \\
    CLIP2Point~\cite{huang2022clip2point} &       & 2.7       & 5.8      & 7.9      & 49.5  & 71.3      & 81.2      & 25.5      & 44.6      & 59.4 \\
     ULIP-PointBERT~\cite{xue2023ulip} &       & 6.2   & 13.6      & 17.9  & 60.4  & 79.0      & 84.4    & 51.5  & 71.1      & 80.2 \\
    OpenShape-SparseConv & & 11.6 & 21.8 & 27.1 & 72.9 & 87.2 & 93.0 & 52.7 & 72.7 & 83.6\\ 
     OpenShape-PointBERT & & 10.8 & 20.2 & 25.0 & 70.3 & 86.9 & 91.3 & 51.3 & 69.4 & 78.4 \\
    \midrule
    ULIP-PointBERT~\cite{liu2023openshape} & \multirow{2}[1]{*}{OpenShape~\cite{liu2023openshape}}  & 21.4  & 38.1      & 46.0      & 71.4  & 84.4      & 89.2      & 46.0     & 66.1      & 76.4 \\
    OpenShape-SparseConv~\cite{liu2023openshape}  &\multirow{2}[1]{*}{(w/o Objaverse-LVIS~\cite{deitke2023objaverse})} & 37.0  &   58.4    & 66.9  & 82.6  &   95.0    & 97.5  &    54.9   &  76.8     &  87.0 \\
     OpenShape-PointBERT~\cite{liu2023openshape} & & 39.1 & 60.8 & 68.9 & \textbf{85.3} & 96.2 & 97.4 & 47.2 & 72.4 & 84.7 \\
    \midrule
    ULIP-PointBERT~\cite{liu2023openshape} & \multirow{3}{*}{OpenShape~\cite{liu2023openshape}} & 26.8  & 44.8      & 52.6      & 75.1  & 88.1      & 93.2      & 51.6      & 72.5      & 82.3 \\
    OpenShape-SparseConv~\cite{liu2023openshape}  &   & 43.4 & 64.8 & 72.4 & 83.4 & 95.6 & 97.8 & \textbf{56.7} &   78.9    & 88.6 \\
    OpenShape-PointBERT~\cite{liu2023openshape} & & \textbf{46.8} & \textbf{69.1} & \textbf{77.0} & 84.4 & \textbf{96.5} & \textbf{98.0} & 52.2 & \textbf{79.7} & \textbf{88.7} \\
    \bottomrule
    \end{tabular}
  \label{tab:OV-3D}
\end{table*}

\begin{table*}[htbp]
    \centering
    \caption{Results for open-world 3D semantic segmentation on ScanNet and nuScenes in terms of hIoU / mIoU$^\mathcal{B}$ / mIoU$^\mathcal{N}$. PLA w/o training denotes training without language supervision in~\cite{ding2023language}.}
    \begin{tabular}{l|c|c|c|c|c}
        \bottomrule[1pt]
        \multirow{2}{*}{Method}  & \multicolumn{3}{c|}{ScanNet~\cite{dai2017scannet}} & \multicolumn{2}{c}{nuScenes~\cite{caesar2020nuscenes}} \\
        \cline{2-6}
        & {B15/N4} & {B12/N7} & {B10/N9} & {B12/N3} & {B10/N5} \\
        \hline
        LSeg-3D~\cite{ding2023language} & 00.0 / 64.4 / 00.0 & 00.9 / 55.7 / 00.1 & 01.8 / 68.4 / 00.9 & 00.6 / 74.4 / 00.3 
        & 0.00 / 71.5 / 0.00 \\
        PLA w/o training~\cite{ding2023language} & 39.7 / 68.3 / 28.0 & 24.5 / \textbf{70.0} / 14.8 & 25.7 / 75.6 / 15.5 & 25.5 / \textbf{75.8} / 15.4 & 10.7 / 76.0 / 05.7 \\
        PLA~\cite{ding2023language} & 65.3 / 68.3 / 62.4 & 55.3 / 69.5 / 45.9 & 53.1 / 76.2 / 40.8 & 47.7 / 73.4 / 35.4 & 24.3 / 73.1 / 14.5\\
        \hline
        RegionPLC~\cite{yang2023regionplc} & \textbf{69.9} / \textbf{68.4} / \textbf{71.5} & \textbf{65.1} / 69.6 / \textbf{61.1} & \textbf{58.8} / \textbf{76.6} / \textbf{47.7} & \textbf{62.0} / \textbf{75.8} / \textbf{52.4} & \textbf{36.6} / \textbf{76.7} / \textbf{24.1} \\
        \hline
        Fully-Sup (Upper Bound)~\cite{yang2023regionplc} & 73.3 / 68.4 / 79.1 & 70.6 / 70.0 / 71.8 & 69.9 / 75.8 / 64.9 & 73.7 / 76.6 / 71.1 & 74.8 / 76.8 / 72.8  \\
        \toprule[0.8pt]
    \end{tabular}
    \label{tab:OV-3D-Seg}
\end{table*}


\section{Benchmark Results}
\label{sec:benchmark_results}

This section systematically compares the different settings and methods in each task. For each setting, first, we introduce the datasets used in the evaluation. Next, we present the details of each setting separately. Then, we compare the results in detailed tables. We list the detailed results and settings for reference.

\begin{table*}[t!]
  \centering
  \caption{\textbf{Open vocabulary object detection results on the COCO dataset.} The evaluation metric is box mAP at IoU threshold 0.5 for novel and base classes, respectively. The best results are emphasized in \textbf{bold}.}
  \begin{adjustbox}{width=0.90\textwidth}
    \begin{tabular}{lcccccccc} 
      \toprule[0.15em]
      Method                                           & \makecell{Vision \\ Backbone}  & \makecell{Text \\ Model}   & \makecell{Trainable \\ Backbone}  & Detector    & Extra Data         & $AP_{novel}$ & $AP_{base}$ & $AP_{all}$ \\
      \toprule[0.15em]
      OVR-CNN~\cite{zareian2021opendet_ovrcnn}    \textit{(CVPR'21)}     & RN50-C4                       & BERT            &     \cmark                         & Faster R-CNN     & COCO Cap        &  22.8 &  46.0 & 39.9 \\
      ViLD-ens~\cite{gu2021open_vild}             \textit{(ICLR '22)}     & RN50-FPN                      & CLIP ViT-B/32   &     \cmark                         & Mask R-CNN      &  -                & 27.6 & 59.5 & 51.3 \\
      RegionCLIP~\cite{Zhong2021RegionCLIPRL}     \textit{(CVPR '22)}     & RN50-C4                       & CLIP RN50       &     \cmark                         & Faster R-CNN     & COCO Cap         & 26.8 & 54.8 & 47.5 \\
      RegionCLIP~\cite{Zhong2021RegionCLIPRL}     \textit{(CVPR '22)}     & RN50-C4                       & CLIP RN50       &     \cmark                         & Faster R-CNN     & CC3M             & 31.4 & 57.1 & 50.4 \\
      Detic~\cite{zhou2022detecting}              \textit{(ECCV '22)}     & RN50-FPN                      & CLIP            &     \cmark                         & CenterNet2      & IN-L             & 27.8 & 47.1 & 45.0\\
      PB-OVD~\cite{Gao2021pbovd}                  \textit{(ECCV '22)}     & RN50-FPN                      & CLIP ViT-B/32   &     \cmark                         & Mask R-CNN       & COCO Cap         &  30.8 & 46.1 & 42.1 \\
      HierKD~\cite{Ma2022HierKD}                  \textit{(CVPR '22)}     & RN50-FPN                      & CLIP            &     \cmark                         & ATSS            & COCO Cap         &  20.3 & 51.3 &43.2 \\
      LocOV~\cite{Bravo2022LocalizedVM}           \textit{(GCPR '22)}     & RN50-FPN                      & CLIP            &     \cmark                         & Faster R-CNN     & COCO Cap         &  28.6 & 51.3 & 45.7 \\
      OV-DETR~\cite{zang2022open}                 \textit{(ECCV '22)}     & RN50-C4                       & CLIP            &     \cmark                         & Deformable DETR & -                & 29.4 & 52.7 & \textbf{61.0} \\
      MEDet~\cite{Chen2022OpenVO}                 \textit{(arXiv'22)}     & RN50-FPN                      & CLIP            &     \cmark                         & Faster R-CNN     & COCO Cap         &  32.6	& 53.5	& 48.0 \\
      PromptDet~\cite{feng2022promptdet}          \textit{(ECCV '22)}     & RN50-FPN                      & CLIP            &     \cmark                         & Mask R-CNN       & LAION          &  31.4 & 29.1 & 29.7 \\
      VLDet~\cite{VLDet}                          \textit{(ICLR '23)}     & RN50-FPN                      & CLIP            &     \cmark                         & CenterNet2      & COCO Cap       & 32.0 & 50.6 & 45.8 \\
      F-VLM~\cite{Kuo2022FVLMOO}                  \textit{(ICLR '23)}     & RN50-FPN                      & CLIP            &     \xmark                         & Mask R-CNN       & -                 & 28.0 & - & 39.6 \\
      BARON~\cite{wu2023baron}                    \textit{(CVPR '23)}     & RN50-FPN                      & CLIP            &     \cmark                         & Faster R-CNN     & -               &  {34.0} & {60.4} & 53.5 \\
      OADP~\cite{OADP}           \textit{(CVPR '23)}     & RN50-C4                       & CLIP            &     \cmark                         & Faster R-CNN     & COCO Cap          &  30.0 & 53.3 & 47.2 \\
      CORA~\cite{wu2023cora}           \textit{(CVPR '23)}     & RN50                       & CLIP            &     \xmark                         & DAB-DETR     & -          &  35.1  & 35.5 & 35.4 \\
      EdaDet~\cite{Shi2023EdaDetOO}   \textit{(ICCV'23)} & RN50 & CLIP-RN50 & \cmark  & DETR & - &  37.8 & 57.7 & 52.5 \\
      CoDet~\cite{ma2023codet}   \textit{(NeurIPS'23)} & RN50 & CLIP & \cmark  & CenterNet2 & COCO Caption &  30.6 & 52.3 & 46.6 \\
      \hline
      RO-ViT~\cite{Ro_ViT}           \textit{(CVPR'23)}     & ViT-B                       & CLIP            &     \xmark                         & Mask R-CNN     & -          &  30.2 & - & 41.5 \\
      DITO~\cite{Kim2023DITO}   \textit{(arXiv'23)} & ViT-B/16 & DITO Pretrain & \cmark  & Faster R-CNN & -&38.6 & - & 48.5 \\
      CLIPSelf~\cite{wu2023clipself}   \textit{(ICLR'24)} & ViT-B/16 & CLIP & \xmark  & Mask R-CNN & - & 37.6& 54.9 & 50.4 \\
      DST-Det~\cite{Xu2023DSTDetSD}   \textit{(arXiv'23)} & ViT-B/16 & CLIP & \xmark  & Mask R-CNN &-& \textbf{41.3} & - &  - \\
      ProxyDet~\cite{Jeong2023ProxyDetSP}   \textit{(AAAI '24)} & RN50-C4 & CLIP  ViT-B/32 & \cmark  & Faster R-CNN & - &  30.4 & 52.6 & 46.8  \\
      LP-OVOD~\cite{Pham2023LPOVODOO}   \textit{(WACV '24)} & RN50 & CLIP   & \cmark  & Faster R-CNN & - & 40.5 & \textbf{60.5} & 55.2  \\
      
      \toprule[0.15em]
    \end{tabular}
  \end{adjustbox}

  \label{table:coco_results}
  \end{table*}
  
\begin{table*}[t!]
  \centering
  \caption{\textbf{Open vocabulary object detection results on LVIS 1.0 dataset.} The evaluation metric is mask mAP, and $AP_r$ represents the performance for novel categories.}
  \begin{adjustbox}{width=0.90\textwidth}
    \begin{tabular}{lccccccccc} 
      \toprule[0.15em]
      Method                                                              & \makecell{Vision \\ Backbone}  & \makecell{Text \\ Model}   & \makecell{Trainable \\ Backbone}  & Detector    & Extra Data          & $AP_{r}$ & $AP_{c}$ & $AP_{f}$ & $AP_{all}$ \\
      \toprule[0.15em]
      ViLD-ens~\cite{gu2021open_vild}          \textit{(ICLR'22)}     & RN50-FPN    & CLIP ViT-B/32   & \cmark  & Mask R-CNN      &  -                & 16.6 & 24.6 & 30.3 & 25.5 \\
      RegionCLIP~\cite{Zhong2021RegionCLIPRL}  \textit{(CVPR'22)}     & RN50-C4     & CLIP RN50       & \cmark  & Faster R-CNN     & CC3M         & 17.1 & 27.4 & 34.0 & 28.2 \\
      Detic~\cite{zhou2022detecting}           \textit{(ECCV'22)}     & RN50-FPN    & CLIP            & \cmark  & CenterNet2      & IN-L             &  19.5 & - & - & 30.9 \\
      OV-DETR~\cite{zang2022open}              \textit{(ECCV'22)}     & RN50-C4     & CLIP            & \cmark  & Deformable DETR & -                 & 17.4 &  25.0 &  32.5 & 26.6\\
      MEDet~\cite{Chen2022OpenVO}              \textit{(arXiv'22)}    & RN50-FPN    & CLIP            & \cmark  & Faster R-CNN     & CC3M       &  22.4	& - & - & 34.4 \\
      DetPro~\cite{Du2022LearningTP} \textit{(ECCV'20)} & RN50-FPN & CLIP ViT-B/32 & \cmark  & Mask R-CNN      &  -                & 19.8 & 25.6 & 28.9 & 25.9 \\
      PromptDet~\cite{feng2022promptdet}       \textit{(ECCV'22)}     & RN50-FPN    & CLIP            & \cmark  & Mask R-CNN       & LAION            &  21.4 & 23.3 & 29.3 & 25.3 \\
      VLDet~\cite{VLDet}                       \textit{(ICLR'23)}     & RN50-FPN    & CLIP            & \cmark  & CenterNet2      & CC3M             & 21.7 & 29.8 & 34.3 & 30.1 \\
      F-VLM~\cite{Kuo2022FVLMOO}               \textit{(ICLR'23)}     & RN50-FPN    & CLIP            & \xmark  & Mask R-CNN       & -                & 18.6 & - & - & 24.2 \\
      BARON~\cite{wu2023baron}                 \textit{(CVPR'23)}     & RN50-FPN    & CLIP            & \cmark  & Faster R-CNN     & -                &  22.6 & 27.6 & 29.8 & 27.6 \\
      OADP~\cite{OADP}        \textit{(CVPR'23)}     & RN50-C4     & CLIP            & \cmark  & Faster R-CNN     & -        & 21.7 & 26.3 & 29.0 & 26.6 \\
      CORA~\cite{wu2023cora}           \textit{(CVPR'23)}     & RN50                       & CLIP            &     \xmark                         & DAB-DETR     & -          & 22.2 & - & - & \\
    EdaDet~\cite{Shi2023EdaDetOO} \textit{(ICCV'23)} & RN50 & CLIP RN50& \cmark & DETR &-& 23.7&27.5&29.1&27.5 \\
      CoDet~\cite{ma2023codet} \textit{(NeurIPS'23)} & RN50 & CLIP& \cmark & CenterNet2 &CC3m & 23.4&30.0&34.6&30.7 \\
    ProxyDet~\cite{Jeong2023ProxyDetSP} \textit{(AAAI '24)} & RN50 & CLIP ViT-B/32& \cmark & Mask R-CNN &IN-L& 26.2&-&-&32.5 \\
      LP-OVOD~\cite{Pham2023LPOVODOO}   \textit{(WACV '24)} & RN50 & CLIP   & \cmark  & Mask R-CNN & - & 19.3 & 29.4 & 26.1 26.2 \\
      \hline
      RO-ViT~\cite{Ro_ViT}           \textit{(CVPR'23)}     & ViT-B                       & CLIP            &     \xmark                         & Mask R-CNN     & -          &  28.0 & - &- & 30.2 \\
      DITO~\cite{Kim2023DITO}   \textit{(arXiv'23)} & ViT-B/16 & DITO Pretrain & \cmark  & Mask R-CNN & 32.5&- & - & -&34.0 \\
      CLIPSelf~\cite{wu2023clipself}   \textit{(ICLR'24)} & ViT-B/16 & CLIP & \xmark  & Mask R-CNN &-& 25.3& 21.8  & 29.1  & 25.2 \\
      DST-Det~\cite{Xu2023DSTDetSD}   \textit{(arXiv'23)} & ViT-B/16 & CLIP & \xmark  & Mask R-CNN &-& 26.2&- & - & -\\
      RegionCLIP~\cite{Zhong2021RegionCLIPRL}  \textit{(CVPR '22)}     & RN50x64-C4  & CLIP RN50x64    & \cmark  & Faster R-CNN     & CC3M            & 22.0 & 32.1 & 36.9 & 32.3 \\
      Detic~\cite{zhou2022detecting} \textit{(ECCV'22)}     & Swin-B      & CLIP & \cmark  & CenterNet2& IN-L    &  23.9 & \textbf{40.2} & \textbf{42.8} & \textbf{38.4} \\
      VLDet~\cite{VLDet} \textit{(ICLR'23)}     & Swin-B    & CLIP            & \cmark  & CenterNet2      & CC3M             & 26.3 & 39.4 &  41.9 & 38.1 \\
      F-VLM~\cite{Kuo2022FVLMOO} \textit{(ICLR'23)}     & RN50x64-FPN    & CLIP            & \xmark  & Mask R-CNN       & -                & {32.8} & - & - & 34.9 \\
     CLIPSelf~\cite{wu2023clipself}   \textit{(ICLR'24)} & ViT-L/16 & CLIP & \xmark  & Mask R-CNN &-&  \textbf{34.9} & 34.6 & 35.6 & 35.1 \\
     DST-Det~\cite{Xu2023DSTDetSD}   \textit{(arXiv'23)} & RN50x64-FPN & CLIP & \xmark  & Mask R-CNN &-& 34.5&- & - & -  \\  
      \toprule[0.15em]
    \end{tabular}
  \end{adjustbox}
  \label{table:lvis_results}
  \end{table*}
  
\begin{table*}
\footnotesize
\centering
  \caption{\textbf{Open vocabulary object detection results on V3Det~\cite{wang2023v3det} dataset.} The metric is mAP on different IoU thresholds.}\vspace{-3mm}
    \begin{adjustbox}{width=0.90\textwidth}
    \begin{tabular}{lcccccccc}
    \toprule[0.15em]
    Method                                                              & \makecell{Vision \\ Backbone}  & \makecell{Text \\ Model}   & \makecell{Trainable \\ Backbone}  & Detector    & Extra Data          & $AP_{novel}$ & $AP_{base}$ & $AP_{all}$ \\
      \toprule[0.15em]
    RegionCLIP~\cite{Zhong2021RegionCLIPRL} \textit{(CVPR'22)}     & RN50    & CLIP            & \cmark  & Faster R-CNN       & -                & 3.1 & 22.1 & 12.6 \\
    Detic~\cite{zhou2022detecting} \textit{(ECCV'22)}     & RN50    & CLIP            & \cmark  & CenterNet2 & -  & 6.7& 30.2 & 17.1 \\
    DST-Det~\cite{Xu2023DSTDetSD} \textit{(arXiv'23)}     & RN50    & CLIP & \xmark  & Faster R-CNN & - & 7.2& - & - \\
    DST-Det~\cite{Xu2023DSTDetSD} \textit{(arXiv'23)}     & RN50x64    & CLIP & \xmark  & Faster R-CNN & - & 13.5 & - & - \\
    \bottomrule[0.15em]
    \end{tabular}
    \end{adjustbox}
\label{table:v3det_results}
\end{table*}

\subsection{Open Vocabulary Object Detection}

\noindent
\textbf{Settings.}
Open vocabulary object detection methods, such as OVR-CNN~\cite{zareian2021opendet_ovrcnn} and ViLD~\cite{gu2021open_vild}, have primarily focused on the COCO~\cite{COCO_dataset}, LVIS~\cite{gupta2019lvis} and V3Det~\cite{wang2023v3det} datasets. 
The COCO dataset comprises 80 classes, with 48 considered base classes and 17 treated as novel classes. 
Any annotations not labeled with base classes are removed from the training data. 
Therefore, for open vocabulary object detection using the COCO dataset, there are 107,761 training images containing 665,387 bounding box annotations of base classes and 4,836 test images with 28,538 bounding box annotations of both base and novel categories. 
On the other hand, the LVIS dataset is designed for long-tail object detection tasks and includes 1203 classes.  866 frequent and common ones serve as base categories, while the remaining rare ones (377) act as novel categories.
The LVIS dataset is specifically designed for long-tail object detection tasks and consists of 1203 classes. Among these, there are 866 frequent and common categories that serve as base categories, while the remaining 377 rare ones serve as novel categories.
The V3Det dataset is a vast vocabulary visual detection dataset that contains 13,204 categories, 243k images, and 1,753k bounding box annotations. In the open vocabulary setting, there are 6,709 base categories and 6,495 novel categories.

\noindent
\textbf{Evaluation Metrics.} When evaluating open vocabulary object detection, we calculate the box mAP using an IoU threshold of 0.5 for the COCO dataset and mask mAP for the LVIS dataset. 
To assess a method's ability to detect novel classes, we split the $AP$ metric into two categories: $AP_{novel}$ and $AP_{base}$. Our main focus is on measuring $AP_{novel}$.

\noindent
\textbf{Results Comparison.}
In Table \ref{table:coco_results}, \ref{table:lvis_results}, and \ref{table:v3det_results}, we present the performance of current open vocabulary object detection methods on COCO, LVIS, and V3Det. BARON~\cite{wu2023baron} outperforms other models on both datasets when using Faster R-CNN detector with the ResNet-50 backbone, achieving a high score of 42.7 $AP_{novel}$ on COCO and 22.6 $AP_{r}$ on LVIS. FVLM~\cite{Kuo2022FVLMOO} achieves 28.0 $AP_{novel}$ on the COCO dataset and 18.6 $AP_{r}$ on LVIS using only frozen backbone and detection data. 
When utilizing the ResNet50x64 backbone, F-VLM achieves a remarkable performance of 32.8 $AP_{r}$ on the LVIS dataset, surpassing other methods that employ the Swin-B backbone and additional data like CC3M and ImageNet. And for one-stage detector methods such as HierKD~\cite{Ma2022HierKD} and GridCLIP~\cite{Lin2023GridCLIPOO}, there is still a significant gap compared to two-stage detectors.

\begin{table*}[t!]
\centering
\caption{\textbf{Open vocabulary semantic segmentation performances} under \textbf{the self-evaluation setting.} The metric is mIoU. ``Harmonic'' means the harmonic mean of the mIoU for base classes and the mIoU for novel classes.}
  \scalebox{0.90}{
  \begin{tabular}{lcc|ccccccccc}
    \toprule[0.15em]
    \multirow{2}{*}{Method} & \multirow{2}{*}{Backbone} & \multirow{2}{*}{VLM} & \multicolumn{3}{c}{COCO-Stuff} & \multicolumn{3}{c}{PASCAL-VOC} & \multicolumn{3}{c}{PASCAL-Context} \\
    & & & Base & Novel & Harmonic & Base & Novel & Harmonic & Base & Novel & Harmonic \\
    \midrule
    ZegFormer~\cite{Decoupling-zero-shot-semantic-segmentation} (\textit{CVPR'22}) & RN-50 & CLIP-ViT-B/16 & 37.4 & 21.4 & 27.2 & - & - & - & - & - & - \\
    PADing~\cite{PADing} (\textit{CVPR'23}) & RN-50 & CLIP-ViT-B/16 & 40.4 & 24.8 & 30.7 & - & - & - & - & - & - \\
    \midrule
    ZegFormer~\cite{Decoupling-zero-shot-semantic-segmentation} (\textit{CVPR'22}) & RN-101 & CLIP-ViT-B/16 & 36.6 & 33.2 & 34.8 & 86.4 & 63.6 & 73.3 & - & - & - \\
    Xu et al.~\cite{simple-baseline-for-ovss} (\textit{ECCV'22}) & RN-101 & CLIP-ViT-B/16 & 39.6 & 43.6 & 41.5 & 79.2 & 78.1 & 79.3 & - & - & - \\
    PADing$^*$~\cite{PADing} (\textit{CVPR'23}) & RN-101 & CLIP-ViT-B/16 & 39.9   & 44.9 & 42.2 & - & - & - & - & - & - \\
    MaskCLIP+~\cite{Denseclip} (\textit{ECCV'22}) & RN-101 & CLIP-RN-50 & 39.6 & \textbf{54.7} & 45.0 & 88.1 & \textbf{86.1} & \textbf{87.4} & \textbf{48.1} & \textbf{66.7} & \textbf{53.3} \\
    FreeSeg~\cite{freeseg} (\textit{CVPR'23}) & RN-101 & CLIP-ViT-B/16 & \textbf{42.2} & 49.1 & \textbf{45.3} & \textbf{91.8} & 82.6 & 86.9 & - & - & - \\
    \bottomrule[0.15em]
  \end{tabular}}
\label{table:results-ovss-self}
\end{table*}

\begin{table*}[t!]
\centering
\caption{\textbf{Open vocabulary semantic segmentation performances} under \textbf{the cross-evaluation setting.} The metric is mIoU. For COCO, different methods use different supervision to train their models, including mask, classification (cls), and caption (cap). PAS-20$^b$ is an evaluation setting using the PASCAL-VOC dataset proposed by OpenSeg~\cite{OpenSeg}. It only assigns the background class in PASCAL-VOC to the pixels having a PC-59 category, which is harder than PAS-20 in general. ``\dag'' means the results are from a re-implementation of CAT-Seg~\cite{cat-seg}. ``*'' means the result is only tested on the novel classes in the class split.}
  \scalebox{0.79}{
  \begin{tabular}{lcc|cccc|cccccc}
    \toprule[0.15em]
    \multirow{2}{*}{Method} & \multirow{2}{*}{Backbone} & \multirow{2}{*}{VLM} & \multicolumn{3}{c}{COCO} & \multirow{2}{*}{Extra data} & \multirow{2}{*}{A-847} & \multirow{2}{*}{PC-459} & \multirow{2}{*}{A-150} & \multirow{2}{*}{PC-59} & \multirow{2}{*}{PAS-20$^b$} & \multirow{2}{*}{PAS-20} \\
    & & & mask & cls & cap & & & & & & & \\
    
    \midrule

    MaskCLIP~\cite{panoptic-MaskCLIP} (\textit{ICML'23}) & RN-50 & CLIP-ViT-L/14 & \cmark & \cmark & \xmark & \textit{None} & 8.2 & 10.0 & 23.7 & 45.9 & - & - \\
    
    \midrule
    ZegFormer\dag~\cite{Decoupling-zero-shot-semantic-segmentation} (\textit{CVPR'22}) & RN-101 & CLIP-ViT-B/16 & \cmark & \cmark & \xmark & \textit{None} & 5.6 & 10.4 & 18.0 & 45.5 & 65.5 & 89.5 \\
    LSeg+~\cite{OpenSeg} (\textit{ECCV'22}) & RN-101 & ALIGN & \cmark & \cmark & \xmark & \textit{None} & 2.5 & 5.2 & 13.0 & 36.0 & 59.0 & - \\
    OpenSeg~\cite{OpenSeg} (\textit{ECCV'22}) & RN-101 & ALIGN & \cmark & \xmark & \cmark & Localized Narrative & 4.4 & 7.9 & 17.5 & 40.1 & 63.8 & - \\
    Xu et al.~\cite{simple-baseline-for-ovss} (\textit{ECCV'22}) & RN-101 & CLIP-ViT-B/16 & \cmark & \cmark & \xmark & \textit{None} & 7.0 & - & 20.5 & 47.7 & & 88.4 \\
    OVSeg~\cite{Mask-adapted-clip} (\textit{CVPR'23}) & RN-101c & CLIP-ViT-B/16 & \cmark & \cmark & \cmark & \textit{None} & 7.1 & 11.0 & 24.8 & 53.3 & - & 92.6 \\
    Han et al.~\cite{global-knowledge-calibration} (\textit{ICCV'23}) & RN-101 & CLIP-RN-50 & \cmark & \cmark & \cmark & \textit{None} & 3.5 & 7.1 & 18.8 & 45.2 & 83.2 & - \\
    CAT-Seg~\cite{cat-seg} (\textit{arXiv'23}) & RN-101 & CLIP-ViT-B/16 & \cmark & \cmark & \xmark & \textit{None} & 8.4 & 16.6 & 27.2 & 57.5 & 78.3 & 93.7 \\
    
    \midrule
    
    GroupViT~\cite{Groupvit} (\textit{CVPR'22}) & ViT-S/16 & \textit{None} & \xmark & \xmark & \xmark & CC12M + YFCC & - & - & - & 22.4 & - & 52.3 \\
    ViL-Seg~\cite{Vil-Seg} (\textit{ECCV'22}) & ViT-B/16 & \textit{None} & \xmark & \xmark & \xmark & CC12M & - & - & - & 16.3* & - & 34.4* \\
    OVSegmentor~\cite{OVSegmentor} (\textit{CVPR'23}) & ViT-B/16 & \textit{None} & \xmark & \xmark & \xmark & CC4M & - & - & - & 20.4 & - & 53.8 \\
    TCL~\cite{TCL} (\textit{CVPR'23}) & ViT-B/16 & CLIP-ViT-B/16 & \xmark & \xmark & \xmark & CC4M + CC12M & - & - & 17.1 & 33.9 & 55.0 & 83.2 \\
    SegCLIP~\cite{SegCLIP} (\textit{ICML'23}) & ViT-B/16 & CLIP-ViT-B/16 & \xmark & \xmark & \cmark & CC3M & - & - & - & 24.7 & - & 52.6 \\
    PACL~\cite{PACL} (\textit{CVPR'23}) & ViT-B/16 & CLIP-ViT-B/16 & \xmark & \xmark & \xmark & CC3M + CC12M + YFCC & - & - & 31.4 & 50.1 & - & 72.3 \\
    PGSeg~\cite{ov-seg-sem-pgseg} (\textit{arXiv'23}) & ViT-S/16 & \textit{None} & \xmark & \xmark & \xmark & CC12M + RedCaps12M & - & - & - & 23.8 & - & 53.2 \\
    PnP-OVSS~\cite{ov-seg-sem-pnpovss} (\textit{arXiv'23}) & ViT-L/14 & BLIP-ViT-L & \xmark & \xmark & \xmark & \textit{None} & - & - & - & 41.9 & - & 55.7 \\
    GEM~\cite{ov-seg-sem-gem} (\textit{arXiv'23}) & ViT-B/16 & CLIP-ViT-B/16 & \xmark & \xmark & \xmark & \textit{None} & - & - & - & 34.5 & - & 46.2 \\
    Self-SEG~\cite{ov-seg-sem-selfguided} (\textit{arXiv'23}) & Focal-T & BLIP-ViT-L & \xmark & \xmark & \xmark & \textit{None} & 6.4 & - & - & - & 41.6 & - \\
    CaR~\cite{ov-seg-sem-clipasrnn} (\textit{arXiv'23}) & ViT-L/14 & CLIP-ViT-B/16 & \xmark & \xmark & \xmark & \textit{None} & - & - & - & 30.5 & - & 67.6 \\
    
    \midrule

    OpenSeg~\cite{OpenSeg} (\textit{ECCV'22}) & Eff-B7 & ALIGN & \cmark & \xmark & \cmark & Localized Narrative & 8.1 & 11.5 & 26.4 & 44.8 & 70.2 & - \\
    ODISE-cap~\cite{ODISE} (\textit{CVPR'23}) & Stable diffusion & CLIP-ViT-L/14 & \cmark & \xmark & \cmark & \textit{None} & 11.0 & 13.8 & 28.7 & 55.3 & 82.7 & - \\
    ODISE~\cite{ODISE} (\textit{CVPR'23}) & Stable diffusion & CLIP-ViT-L/14 & \cmark & \cmark & \xmark & \textit{None} & 11.1 & 14.5 & 29.9 & 57.3 & \textbf{84.6} & - \\
    OVSeg~\cite{Mask-adapted-clip} (\textit{CVPR'23}) & Swin-B & CLIP-ViT-L/14 & \cmark & \cmark & \cmark & \textit{None} & 9.0 & 12.4 & 29.6 & 55.7 & - & 94.5 \\
    SAN~\cite{side-adapter} (\textit{CVPR'23}) & ViT-L/14 & CLIP-ViT-L/14 & \cmark & \cmark & \xmark & \textit{None} & 12.4 & 15.7 & 32.1 & 57.7 & - & 94.6 \\
    CAT-Seg~\cite{cat-seg} (\textit{arXiv'23}) & Swin-B & CLIP-ViT-L/14 & \cmark & \cmark & \xmark & \textit{None} & 10.8 & 20.4 & 31.5 & 62.0 & 81.8 & 96.6 \\
    SED~\cite{ov-seg-sem-sed} (\textit{arXiv'23}) & - & CLIP-ConvNeXt-L & \cmark & \cmark & \xmark & \textit{None} & 13.9 & \textbf{22.6} & \textbf{35.2} & 60.6 & - & 96.1 \\
    SCAN~\cite{cat-seg} (\textit{arXiv'23}) & Swin-B & CLIP-ViT-L/14 & \cmark & \cmark & \xmark & \textit{None} & \textbf{14.0} & 16.7 & 33.5 & 59.3 & - & 97.2 \\

    \midrule

    X-Decoder~\cite{X-decoder} (\textit{CVPR'23}) & DaViT-L & \textit{None} & \cmark & \cmark & \cmark & \cite{conceptual-captions, SBU, visual-genome, coco-captions} & 9.2 & 16.1 & 29.6 & \textbf{64.0} & - & \textbf{97.7} \\
    OpenSeed~\cite{OpenSeeD} (\textit{ICCV'23}) & Swin-L & \textit{None} & \cmark & \cmark & \xmark & Objects365~\cite{objects365} & - & - & 23.4 & - & - & - \\
    HIPIE~\cite{HIPIE} (\textit{arXiv'23}) & ViT-H & \textit{None} & \cmark & \cmark & \xmark & Objects365~\cite{objects365} & - & - & 29.0 & - & - & - \\
    UOVN~\cite{UOVN} (\textit{arXiv'23}) & RN-50 & \textit{None} & \cmark & \cmark & \xmark & \cite{refcoco, referitgame, visual-genome, phrasecut} & 13.5 & 17.1 & 30.7 & 54.3 & - & - \\
    
    \bottomrule[0.15em]
  \end{tabular}}
\label{table:results-ovss-cross}
\end{table*}

\subsection{Open Vocabulary Semantic Segmentation}

\noindent
\textbf{Settings.}
There are two main settings in open vocabulary semantic segmentation, according to different train/test data choices. \textit{\textbf{1) Self-evaluation setting.}} Methods like ZegFormer~\cite{Decoupling-zero-shot-semantic-segmentation} and MaskCLIP+~\cite{Denseclip} train and test their models on the train/test split from the same dataset. For example, ZegFormer trains on base annotations on the COCO-Stuff training set and tests on the COCO-Stuff validation set with base and novel categories. \textit{\textbf{2) Cross-evaluation setting.}} Methods like OpenSeg~\cite{OpenSeg} and OVSeg~\cite{Mask-adapted-clip} train their models on COCO-Stuff and test on other datasets, such as ADE20K, Pascal VOC, and Pascal Context. There is also another setting that splits a dataset into multiple folds and performs n-fold evaluations. We do not compare the results under this setting considering only few works adopt it~\cite{Language-driven-semantic-segmentation,Fusioner}.

The most commonly used datasets for the cross-evaluation setting are ADE20K, Pascal Context, and Pascal VOC. The ADE20K dataset has 2k validation images that comprise various indoor and outdoor semantic categories. The full dataset contains 2,693 foreground and background classes. Open vocabulary works often evaluate their models with the 847 classes split, named A-847, and the 150 classes split, named A-150. The Pascal Context dataset has 5k validation images with 459 categories (PC-459). Some works also evaluate their models on the dataset with the 59 most frequent classes (PC-59). The Pascal VOC dataset has 20 classes used for evaluation, denoted as PAS-20. OpenSeg~\cite{OpenSeg} proposes only assigning pixels with a Pascal Context category to background classes. We denote this setting as PAS-20$^b$. 

\noindent
\textbf{Evaluation Metrics.}
We adopt mean Intersection-over-Union (mIoU) to evaluate a model's performance on open vocabulary semantic segmentation. Specifically, for the cross-evaluation setting, the IoU averaged on all the categories in the evaluation datasets is listed because all the classes can be seen as novel. For the self-evaluation setting, the testing dataset usually contains both base and novel classes. For a fair comparison, we only report mIoUs on novel classes while highlighting that the number is in a different setting.

\noindent
\textbf{Results under the Self-evaluation Setting.}
Tab.~\ref{table:results-ovss-self} shows the open vocabulary semantic segmentation results under the self-evaluation setting. MaskCLIP+~\cite{Denseclip} achieves the highest novel mIoU on COCO-Stuff, PASCAL-VOC, and PASCAL-Context, while in general, FreeSeg~\cite{freeseg} has the best scores on base and harmonic mIoU. PADing$^*$ means including complicated crop-mask image preprocess (CLIP-Image encoder).

\noindent
\textbf{Results under the Cross-evaluation Setting.}
Tab.~\ref{table:results-ovss-cross} shows the open vocabulary semantic segmentation results under the cross-evaluation setting. SCAN~\cite{dai2017scannet} achieves the highest mIoU of 14.0 on the A-847 dataset. SED~\cite{ov-seg-sem-sed} achieves 22.6 on the PC-459 dataset and 35.2 on the A-150 dataset. X-Decoder~\cite{X-decoder} achieves 65.1 and 97.9 on PC-59 and PAS-20 datasets. For the PAS-20$^b$ dataset, ODISE~\cite{ODISE} achieves the highest mIoU of 84.6. Note that the works at the bottom blank use large-scale extra data for training. For example, X-Decoder~\cite{X-decoder} uses Conceptual Captions~\cite{conceptual-captions}, SBU Captions~\cite{SBU}, Visual Genome~\cite{visual-genome}, and COCO Captions~\cite{coco-captions} as its training set. It contains 4M image-text pairs. Among works training without pixel-level annotations, PACL~\cite{PACL} achieves a much higher score than others. It uses the biggest extra training data that contains CC3M, CC12M, and YFCC.

\begin{table*}[t!]
   \centering
    \caption{\textbf{Open vocabulary instance segmentation} performances on the COCO-Instances dataset.} \vspace{-2mm}
   \scalebox{0.9}{
   \begin{tabular}{lcc|cccc|cc|ccc}
      \toprule[0.15em]
      \multirow{2}{*}{Method} & \multirow{2}{*}{Backbone} & \multirow{2}{*}{VLM} & \multicolumn{3}{c}{COCO} & \multirow{2}{*}{Extra data} & \multicolumn{2}{c|}{Constrained}  & \multicolumn{3}{c}{Generalized} \\
        & & & mask & cls & cap & & $AP_{base}$ & $AP_{novel}$ & $AP_{base}$ & $AP_{novel}$ & $AP_{all}$ \\
        \hline
        D$^2$Zero~\cite{D2Zero} (\textit{CVPR'23}) & RN-50 & CLIP-ViT-B/16 & \cmark & \cmark & \xmark & \textit{None} & - & 23.7 & 40.9 & 21.9 & -\\
        XPM~\cite{XPM} (\textit{CVPR'22}) & RN-50 & \textit{None} & \cmark & \cmark & \cmark & CC3M & 42.4 & 24.0 & 41.5 & 21.6 & 36.3 \\
        Mask-free OVIS~\cite{mask-free-OVIS} (\textit{CVPR'23}) & RN-50 & ALBEF & \xmark & \xmark & \cmark & \textit{None} & - & 27.4 & - & 25.0 & - \\
        CGG~\cite{CGG} (\textit{ICCV'23}) & RN-50 & \textit{None} & \cmark & \cmark & \cmark & \textit{None} & \textbf{46.8} & \textbf{29.5} & \textbf{46.0} & \textbf{28.4} & \textbf{41.4} \\
      \bottomrule[0.10em]
   \end{tabular}}
   \label{tab:result_OVIS}
\end{table*}

\begin{table*}[t!]
   \centering
    \caption{\textbf{Open vocabulary panoptic segmentation} performances on the ADE20K dataset. No methods use extra data for training.}\vspace{-3mm} 
   \scalebox{1.0}{
   \begin{tabular}{lcc|ccc|ccccc}
      \toprule[0.15em]
      \multirow{2}{*}{Method} & \multirow{2}{*}{Backbone} & \multirow{2}{*}{VLM} & \multicolumn{3}{c|}{COCO} & \multirow{2}{*}{PQ} & \multirow{2}{*}{PQ$^{th}$} & \multirow{2}{*}{PQ$^{st}$} & \multirow{2}{*}{SQ} & \multirow{2}{*}{RQ} \\
        & & & mask & cls & cap & & & & & \\
        \hline
        MaskCLIP~\cite{panoptic-MaskCLIP} (\textit{ICML'23}) & RN-50 & CLIP-ViT-L/14 & \cmark & \cmark & \xmark & 15.1 & 13.5 & 18.3 & \textbf{70.5} & 19.2 \\
        ODISE~\cite{ODISE} (\textit{CVPR'23}) & Stable diffusion & CLIP-ViT-L/14 & \cmark & \cmark & \xmark & 22.6 & - & - & - & - \\
        ODISE-cap~\cite{ODISE} (\textit{CVPR'23}) & Stable diffusion & CLIP-ViT-L/14 & \cmark & \xmark & \cmark & \textbf{23.4} & - & - & - & - \   \\
        OPSNet~\cite{OPSNet} (\textit{ICCV'23}) & Swin-L & CLIP-RN-50 & \cmark & \cmark & \xmark & 17.7 & \textbf{15.6} & \textbf{21.9} & 54.9 & \textbf{21.6} \\
      \bottomrule[0.10em]
   \end{tabular}}
   \label{tab:result_OVPS_ade}
\end{table*}

\begin{table*}[t!]
   \centering
    \caption{\textbf{Open vocabulary panoptic segmentation} performances on the COCO dataset. No methods use extra data for training.}
   \scalebox{1.0}{
   \begin{tabular}{lcc|ccc|cccccc}
      \toprule[0.15em]
      \multirow{2}{*}{Method} & \multirow{2}{*}{Backbone} & \multirow{2}{*}{VLM} & \multicolumn{3}{c|}{COCO} & \multirow{2}{*}{PQ$^{s}$} & \multirow{2}{*}{SQ$^{s}$} & \multirow{2}{*}{RQ$^{s}$} & \multirow{2}{*}{PQ$^{u}$} & \multirow{2}{*}{SQ$^{u}$} & \multirow{2}{*}{RQ$^{u}$}\\
        & & & mask & cls & cap & & & & & & \\
        \hline
        PADing~\cite{PADing} (\textit{CVPR'23}) & RN-50 & CLIP-ViT-B/16 & \cmark & \cmark & \xmark & \textbf{41.5} & \textbf{80.6} & \textbf{49.7} & 15.3 &72.8 &18.4 \\
        Freeseg~\cite{freeseg} (\textit{CVPR'23}) & RN-101 & CLIP-ViT-B/16 & \cmark & \cmark & \xmark & 31.4 & 78.3 & 38.9 & \textbf{29.8} & \textbf{79.2} & \textbf{37.6} \\
        
      \bottomrule[0.10em]
   \end{tabular}}
   \label{tab:result_OVPS_coco}
\end{table*}

\subsection{Open Vocabulary Instance Segmentation}

\noindent
\textbf{Settings.}
Following previous works~\cite{zareian2021opendet_ovrcnn, CGG, XPM}, we adopt two settings in evaluating open vocabulary instance segmentation. \textbf{1) Constrained setting.} The base and novel categories are evaluated separately. \textbf{2) Generalized setting.} The base and novel categories are evaluated together. Concretely, in the generalized setting, all the category names are input to the model. The model needs to not only classify novel classes but also to distinguish novel classes from base ones. The second setting is harder than the first one.

\noindent
\textbf{Evaluation Metrics}
We adopt mask mean Average Precision (mAP) as the evaluation metric. For the constrained setting, we report $AP_{base}$ and $AP_{novel}$ separately. For the generalized setting, we report mAP for the base, novel, and all classes.

\noindent
\textbf{Results on COCO.}
Tab.~\ref{tab:result_OVIS} shows the open vocabulary instance segmentation results on the COCO dataset. The CGG method~\cite{CGG} achieves the best results on both constrained and generalized settings while not using pre-trained VLMs or any extra data. And Mask-free OVIS~\cite{mask-free-OVIS} gets a relatively high score on novel classes without any mask labels.

\subsection{Open Vocabulary Panoptic Segmentation}

\noindent
\textbf{Settings.}
The models are trained using COCO-Panoptic, and tested on other datasets in a zero-shot manner. We report both results evaluated on the ADE20K and COCO datasets.

\noindent
\textbf{Evaluation Metrics}
Following previous works~\cite{CGG,PADing,embedding-modulation,ODISE}, we mainly adopt Panoptic Quality (PQ), Segmentation Quality (SQ), and Recognition Quality (RQ) as the evaluation metrics.

\noindent
\textbf{Results.}
Tab.~\ref{tab:result_OVPS_ade} shows the results on the ADE20K dataset. ODISE-cap~\cite{ODISE} achieves the best PQ score of 23.4. It surpasses the second-best score by 0.8. Tab.~\ref{tab:result_OVPS_coco} shows the results on the COCO dataset. PADing~~\cite{PADing} achieves a better PQ of 41.5 for seen classes, while Freeseg~\cite{freeseg} achieves the highest PQ score of 29.8 for unseen classes.

\subsection{Open Vocabulary Video Recognition}
\noindent
\textbf{Settings.}
In video classification, existing methods usually test the zero-shot capability on downstream datasets (e.g., UCF and HMDB) pre-trained on Kinetics-400~\cite{carreira2017quo} datasets. In most of the existing methods such as~\cite{rasheed2022fine,kahatapitiya2023victr}, the ground truth of the Kinetics-400 is used for training, while recent method MAXI~\cite{lin2023match} does not require the ground truth when training on Kinetics-400.

\noindent
\textbf{Results Comparison.} The comparison results are in Tab.\ref{table:OV-VC}. Among the existing methods, Open-VCLIP~\cite{weng2023transforming} performs best on all three downstream datasets. It is also worth noting that the MAXI~\cite{lin2023match} performs well even without annotations from the Kinetics-400.

\subsection{Open Vocabulary Video Instance Segmentation}
\noindent
\textbf{Settings.} To test the performance of open vocabulary video instance segmentation methods, MindVLT~\cite{wang2023towards} collects LV-VIS datasets inheriting the categories of LVIS~\cite{gupta2019lvis} and splits the categories into 659 base categories (frequent and common) and 553 novel categories. In the evaluation protocol, the LV-VIS is not used for training but for evaluation only. The mean Average Precision (mAP) is reported for comparison. $mAP_{base}$ and $mAP_{novel}$ refer to mAP for base and novel categories, respectively.

\noindent
\textbf{Results Comparison.} The comparison results are in Tab.\ref{tab:OV-VIS}. Among the methods, MindVLT~\cite{wang2023towards} achieves the best results on both base and novel categories. MindVLT~\cite{wang2023towards} shows stronger improvement in the novel categories.

\subsection{Open Vocabulary 3D Recognition}
\noindent
\textbf{Settings.}
For open vocabulary 3D recognition, the zero-shot classification results on three different scale datasets, including ModelNet40~\cite{wu20153d}, ScanObjectNN~\cite{uy2019revisiting}, and Objaverse-LVIS~\cite{deitke2023objaverse} are reported. The first two datasets have 40 and 15 common categories, and the Objaverse-LVIS has 1156 LVIS~\cite{gupta2019lvis} categories. The methods are trained on different datasets but tested on the three datasets for fair comparison.

\noindent
\textbf{Results Comparison.} 
As in Tab.\ref{tab:OV-3D}, using the large-scale OpenShape~\cite{liu2023openshape} datasets and larger backbones can significantly boost the performance on the downstream testing datasets.

\subsection{Open Vocabulary 3D semantic segmentation}
\noindent
\textbf{Settings.}
To test the open vocabulary 3D semantic segmentation methods, two datasets, ScanNet~\cite{dai2017scannet} and nuScenes~\cite{caesar2020nuscenes}, are adopted for covering broad application scenarios. The ScanNet and nuScenes have 19 and 15 categories, respectively. The categories are split into base and novel categories for testing. For example, B15/N4 refers to 15 base categories for training and four novel categories that are missing in the training set. The mIoU$^\mathcal{B}$ (base category mIoU), mIoU$^\mathcal{N}$ (novel category mIoU), and hIoU (harmonic mIoU) are reported for comparison.

\noindent
\textbf{Results Comparison.} 
As in Tab.~\ref{tab:OV-3D-Seg}, RegionPLC~\cite{yang2023regionplc} achieves remarkable performance on novel categories and thus indicates that it has a good capability to generalize to unseen categories.

\ifCLASSOPTIONcaptionsoff
  \newpage
\fi



{
\bibliographystyle{IEEEtran}
\bibliography{IEEEabrv,egbib}
}

\end{document}